\documentclass[letterpaper]{article} 
\usepackage{aaai25}
\usepackage{times}  
\usepackage{helvet}  
\usepackage{courier}  
\usepackage[hyphens]{url}  
\usepackage{graphicx} 
\usepackage{natbib}  
\usepackage{caption} 
\usepackage{algorithm}
\usepackage{algorithmic}

\usepackage{tabularx}
\usepackage[table]{xcolor}
\usepackage{amsmath}
\usepackage{booktabs}
\usepackage{multirow}
\usepackage{colortbl}
\usepackage{pifont}
\usepackage{algorithm}
\usepackage{algorithmic}
\usepackage{xcolor}
\definecolor{darkgreen}{rgb}{0.0, 0.6, 0.0} 
\definecolor{darkred}{rgb}{0.7, 0.0, 0.0}   

\usepackage{newfloat}
\usepackage{listings}
\DeclareCaptionStyle{ruled}{labelfont=normalfont,labelsep=colon,strut=off} 
\floatstyle{ruled}
\newfloat{listing}{tb}{lst}{}
\floatname{listing}{Listing}
\nocopyright 

\title{Filling Memory Gaps: Enhancing Continual Semantic Parsing via SQL Syntax Variance-Guided LLMs without Real Data Replay}
\author{
    Ruiheng Liu\textsuperscript{\rm 1, 2},
    Jinyu Zhang\textsuperscript{\rm 2},
    Yanqi Song\textsuperscript{\rm 2},
    Yu Zhang\textsuperscript{\rm 2}\footnotemark[1],
    Bailong Yang\textsuperscript{\rm 1}\footnotemark[1]
}
\affiliations{
    \textsuperscript{\rm 1}Xi’an Research Institute of High-Tech, Xi’an, China\\
    \textsuperscript{\rm 2}Harbin Institute of Technology, Harbin, China\\

    \{rhliu, jyzhang, yqsong, zhangyu\}@ir.hit.edu.cn, xa\_403@163.com
}

\begin{document}

\maketitle

\renewcommand{\thefootnote}{\fnsymbol{footnote}}
\footnotetext[1]{Corresponding Authors.}
\renewcommand{\thefootnote}{\arabic{footnote}}

\begin{abstract}
Continual Semantic Parsing (CSP) aims to train parsers to convert natural language questions into SQL across tasks with limited annotated examples, adapting to the real-world scenario of dynamically updated databases. Previous studies mitigate this challenge by replaying historical data or employing parameter-efficient tuning (PET), but they often violate data privacy or rely on ideal continual learning settings. To address these problems, we propose a new \textbf{L}arge Language Model (LLM)-\textbf{E}nhanced \textbf{C}ontinuous \textbf{S}emantic \textbf{P}arsing method, named \textsc{Lecsp}, which alleviates forgetting while encouraging generalization, without requiring real data replay or ideal settings. Specifically, it first analyzes the commonalities and differences between tasks from the SQL syntax perspective to guide LLMs in reconstructing key memories and improving memory accuracy through a calibration strategy. Then, it uses a task-aware dual-teacher distillation framework to promote the accumulation and transfer of knowledge during sequential training.
Experimental results on two CSP benchmarks show that our method significantly outperforms existing methods, even those utilizing data replay or ideal settings. Additionally, we achieve generalization performance beyond the upper limits, better adapting to unseen tasks.

\end{abstract}

%

\section{Introduction}
Semantic parsing provides a convenient data querying interface for non-expert users to conduct various data analyses \cite{hu2023chatdb,NEURIPS2023_83fc8fab}.
Most previous studies focus on scenarios with static data distributions \cite{10.1609/aaai.v37i11.26535,wang2024improving, 10.1007/978-981-99-8076-5_25}. With the frequent updates of databases in the real world, researchers turn their attention to continual semantic parsing \cite{li-etal-2021-total,lialin2021update,yadav-etal-2023-exploring,10.1609/aaai.v37i11.26492,NEURIPS2023_398b00a0}. 

\begin{figure}[t]
  \includegraphics[width=\columnwidth]{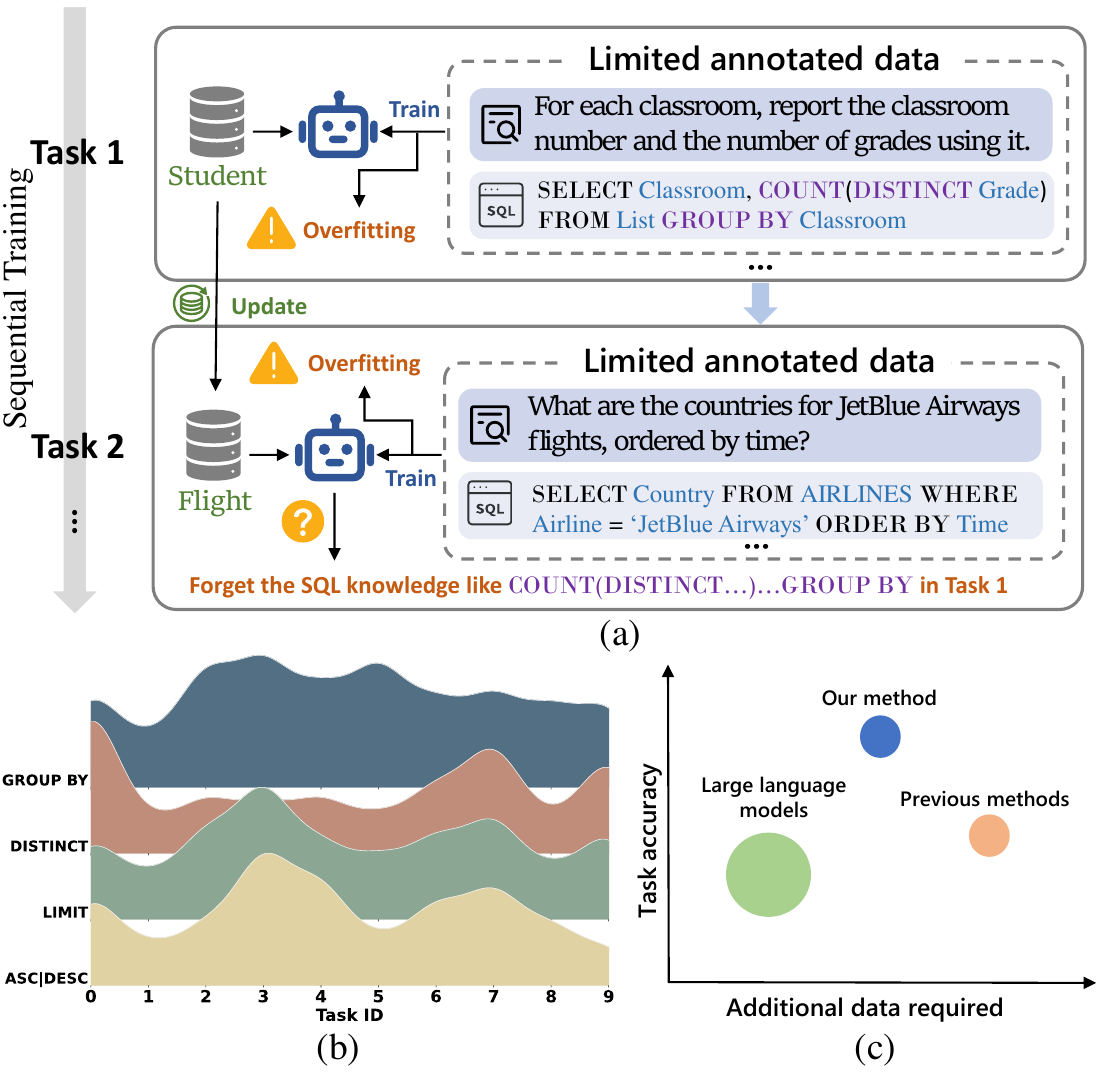}
  \caption{(a) Description of the CSP. (b) The average number of single SQL keywords at cluster centers of different tasks (Section: Inter-Task Memory Completion) in Spider-stream-semi dataset \cite{10.1609/aaai.v37i11.26492}. (c) Comparison between our method and others, where \emph{additional data required} refers to extra historical data or unsupervised data.}
  \label{fig1}
\end{figure}

As illustrated in Figure~\ref{fig1}(a), a semantic parser must undergo continual training across a series of tasks from different databases, while ensuring good performance on current and historical tasks.
This task faces two primary challenges: 
(1) The scarcity of annotated data for each task can easily result in model overfitting \cite{10.1609/aaai.v37i11.26492,qin2022lfpt}.
(2) Training sequentially leads to catastrophic forgetting \cite{10.1609/aaai.v37i11.26492,NEURIPS2023_398b00a0,10444954}, where the model's performance on earlier tasks significantly worsens after learning new ones.

To tackle these challenges, several studies have explored continual learning methods, which can be divided into two types from data and model dimensions: \emph{Rehearsal-based} and \emph{PET-based}. The former devises specific sampling strategies to replay historical task data to prevent forgetting \cite{li-etal-2021-total,10.1609/aaai.v37i11.26492,wang2022memory}. The latter freezes the model backbone and utilizes a limited number of parameters such as prompt \cite{lester-etal-2021-power} to steer representation learning \cite{NEURIPS2023_398b00a0, razdaibiedina2023progressive, yadav-etal-2023-exploring}. 
Although these methods have made initial progress, they still face limitations.
\textbf{First}, the performance of \emph{Rehearsal-based} methods relies heavily on the amount of historical data and may not be practical in privacy-sensitive or memory-constrained environments \cite{wang-etal-2023-rehearsal,10444954,10377536}.
\textbf{Second}, \emph{PET-based} methods learn PET modules for each task and store them in a cache pool. When test samples arrive, most methods assume that the corresponding PET module for each sample is known. This ideal setting contradicts real-world scenarios and hinders generalization to unseen samples \cite{NEURIPS2023_398b00a0, razdaibiedina2023progressive}.

Considering the above, we believe an ideal strategy should not rely on 
any historical data or an ideal continual learning setup. Instead, it should effectively bridge the gaps between tasks by identifying and analyzing the underlying commonalities and critical differences in the knowledge they require, thereby mitigating forgetting and promoting generalization. 
As shown in Figure~\ref{fig1}(b), limited annotated data can lead to uneven distribution of SQL syntax knowledge among tasks. 
Our motivation is to leverage these variances to guide LLMs in reconstructing relevant memories to fill these gaps.

In this work, we propose a novel \textbf{L}LM-\textbf{E}nhanced \textbf{C}ontinual \textbf{S}emantic \textbf{P}arsing method (\textsc{Lecsp}). 
It firstly analyzes component biases from the perspective of SQL syntax between current and past tasks without accessing historical data. Then, these biases are used to guide LLMs in generating relevant pseudo-samples as memory from intra-task and inter-task perspectives, reinforcing commonalities and filling in differences. Additionally, a calibration strategy is introduced to enhance the accuracy and fidelity of memory. To further improve memory utilization efficiency, a task-aware dual-teacher distillation framework is designed. This framework promotes knowledge transfer and accumulation across tasks while facilitating the migration of LLMs capabilities to smaller models. A comparison of our approach with others is illustrated in Figure~\ref{fig1}(c).
Extensive experiments on the CSP benchmarks Spider-stream-semi \cite{10.1609/aaai.v37i11.26492} and Combined-stream \cite{NEURIPS2023_398b00a0} demonstrate that \textsc{Lecsp} significantly outperforms other baselines that use data replay or ideal continual learning setups, with gains up to 8.8\%, and further expanding to 11.4\% in more challenging scenarios. Additionally, we achieve performance beyond theoretical upper bounds in knowledge forward transfer capability. The main contributions of this work are summarized as follows:
\begin{itemize}
    \item We propose a novel CSP framework \textsc{Lecsp}, including memory reconstruction with LLMs and task-aware dual-teacher distillation learning. These help to mitigate forgetting and promote generalization. 
    \item A memory calibration strategy is introduced, consisting of iterative self-correction and SQL skeleton-based sampling, to further improve memory accuracy and fidelity.
    \item Extensive experiments on benchmark datasets show that \textsc{Lecsp} achieves state-of-the-art (SOTA) performance without using historical data or ideal continual learning setups, and surpasses the upper bounds in knowledge transfer capabilities.
\end{itemize}

\section{Related Work}
\subsection{Semantic Parsing with LLMs}
The purpose of semantic parsing is to simplify data access in relational databases for users.
Most previous research \cite{10.1609/aaai.v37i11.26535,cai-etal-2022-star,Dou2023} relies on sequence-to-sequence architectures needing fine-tuning on large datasets. However, SQL annotation of natural language is costly and requires domain-specific experts.
Recently, with the rise of LLMs like GPT-4 \cite{openai2024gpt4} and PaLM-2 \cite{anil2023palm}, some methods have leveraged in-context learning to achieve SOTA performance and reduce labeled data needs \cite{NEURIPS2023_72223cc6,NEURIPS2023_83fc8fab,mao-etal-2024-enhancing}. However, relying on closed-source LLMs poses database security and privacy risks \cite{xue2024dbgpt}, incurs high inference costs, and hinders continual learning due to their black-box nature \cite{10.1145/3654930}.
Unlike previous work, we leverage LLMs to guide smaller models in dynamically adapting to different tasks in CSP and facilitate capability transfer. Notably, to better meet security and cost requirements in real-world applications, we focus on open-source LLMs that can be deployed offline.

\subsection{Continual Learning}
A significant feature of human intelligence is the ability to learn new tasks while accumulating experience and preventing the forgetting of old knowledge, a concept known as continual learning \cite{NEURIPS2019_f8d2e80c,Wang2023}. There are two main challenges: (1) Preventing catastrophic forgetting. (2) Enabling forward transfer of knowledge from past to new tasks.
Previous research on continual learning has primarily focused on classification tasks \cite{han-etal-2020-continual,razdaibiedina2023progressive,9577337}. Recently, with advancements in Pre-trained Language Models (PLMs), it has also garnered widespread attention in complex tasks such as generative tasks \cite{10.1609/aaai.v37i11.26492,NEURIPS2023_398b00a0,yadav-etal-2023-exploring, zhao2024saptsharedattentionframework, liang-etal-2024-continual}. 
\textsc{SFNET} \cite{10.1609/aaai.v37i11.26492} adopts a semi-supervised learning approach, expanding data through self-training on additional unsupervised data. It also designs specific strategies for replaying historical instances to mitigate catastrophic forgetting.
\textsc{C3} \cite{NEURIPS2023_398b00a0} freezes the core parameters of the PLM while training small-scale prompts and performing in-context tuning with relevant instances. In an ideal setting, this approach can prevent forgetting issues.
In this study, we focus on more realistic and challenging CSP scenarios, without accessing any historical data or relying on ideal task settings.

\begin{figure*}[ht]
\centering
\includegraphics[width=\textwidth]{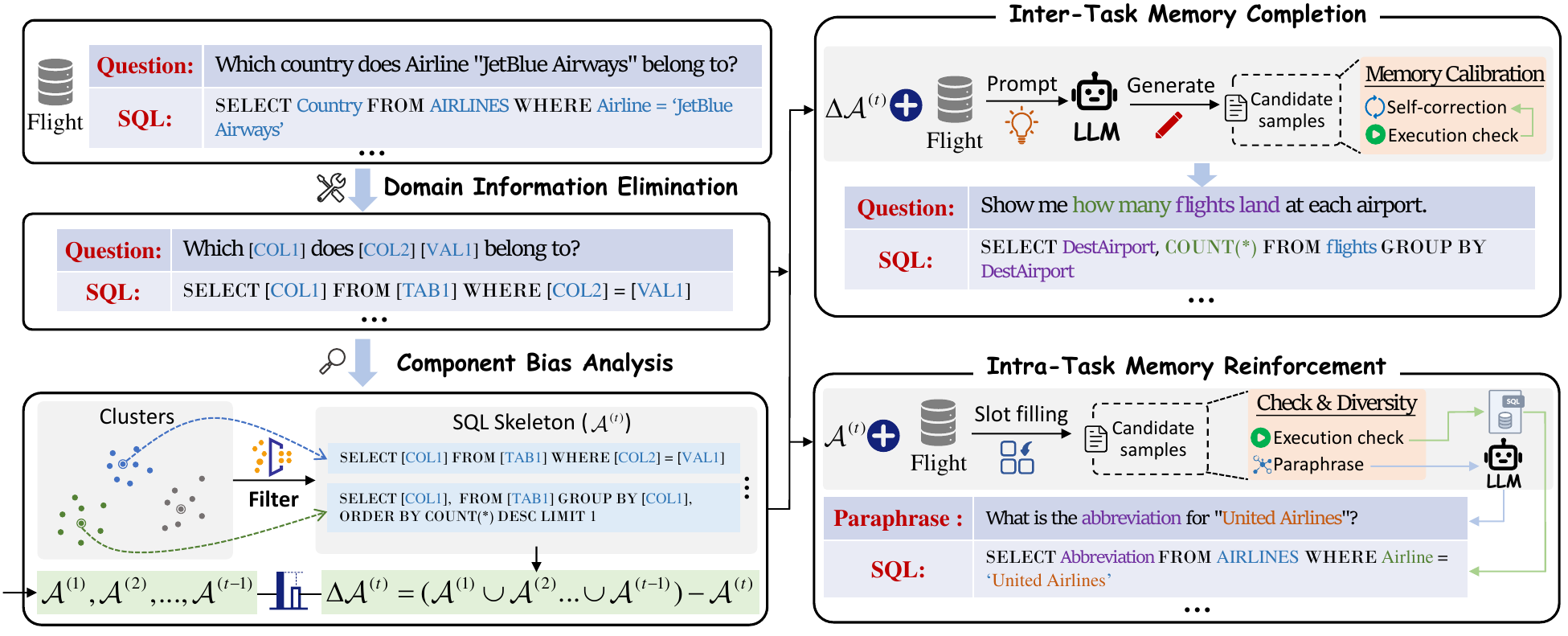}
\caption{Memory reconstruction on task $t$ via LLMs. First, domain information in the questions and SQL pairs are removed, which means replacing entity-link results uniformly with symbols such as $\texttt{[COL]}$ and $\texttt{[VAL]}$. Next, $K$-means clustering is used to obtain the set of SQL skeleton $\mathcal{A}^{(t)}$ for task $t$, and the component bias $\Delta\mathcal{A}^{(t)}$ is derived by taking the difference with the saved sets from previous tasks $\mathcal{A}^{(1)},...,\mathcal{A}^{(t-1)}$. Note that the set $\mathcal{A}$ is only related to SQL syntax and does not involve any historical data. Finally, $\mathcal{A}^{(t)}$ and $\Delta\mathcal{A}^{(t)}$ are used to guide intra-task and inter-task memory reconstruction, respectively reinforcing commonalities and filling differences between tasks.
} 
\label{fig:data_gene}
\end{figure*}

\section{Preliminaries}
Given a natural language question $\mathcal{Q}$ and a database schema $\mathcal{S}=(\mathcal{C}, \mathcal{T})$ consisting of columns $\mathcal{C}=\{c_1^{t_1},c_2^{t_1},...,c_1^{t_2},c_2^{t_2},...\}$ and tables $\mathcal{T}=\{t_i\}_{i=1}^{|\mathcal{T}|}$. The goal of the semantic parsing task is to generate the corresponding SQL query $\mathcal{Y}=\mathcal{F}_{\theta}(\mathcal{Q},\mathcal{S})$, where $\mathcal{F}_{\theta}$ is the parser with parameters $\theta$, and $\mathcal{C}$ and $\mathcal{T}$ are respectively the sets of column names and table names, with $c_i^{t_j}$ indicating the $i$-th column in table $t_j$.
Following the problem definition of previous works \cite{li-etal-2021-total,lialin2021update,  10.1609/aaai.v37i11.26492,NEURIPS2023_398b00a0}, we assume that the semantic parser in CSP is not confined to a single specific train and test set. Instead, it is required to handle a series of semantic parsing tasks $\{\mathcal{D}^1,\mathcal{D}^2,...,\mathcal{D}^M\}$, referred to as a task stream. Each task $\mathcal{D}^i$ originates from a different database domain, formally, for ${\forall\mathcal{D}^i}$ and ${\forall\mathcal{D}^j}$, if $i\neq j$, then $\mathcal{S}(\mathcal{D}^i) \cap \mathcal{S}(\mathcal{D}^j)=\emptyset$, where $\mathcal{S}(\mathcal{D}^i)$ represents the set of all database schemas $\mathcal{S}$ in task $\mathcal{D}^i$. Additionally, each task $\mathcal{D}^i$ has its own training, validation, and test sets.

The primary objective of CSP is to develop a semantic parser that not only performs excellently on previous tasks but also effectively generalizes to future unseen tasks after being sequentially trained on all tasks.

\section{Methodology}
Our method can be simply divided into two stages: {memory reconstruction} and {dual distillation learning}. The first stage is shown in Figure~\ref{fig:data_gene}, we remove domain-specific information from individual tasks and then perform component bias analysis in conjunction with SQL syntax to identify the main characteristics of the current task and its differences from historical tasks. 
Finally, we construct relevant memories from both intra-task and inter-task. In the second stage, as shown in Figure~\ref{fig:framework}, we enhance the efficient utilization of these memories in the continual learning process through a designed task-aware dual-teacher distillation learning framework.

\subsection{Reconstruct Memory with LLMs}
\label{sec:data_gen}

\subsubsection{Inter-Task Memory Completion}
\label{sec:ood}

\paragraph{Domain Information Elimination} 
To mitigate the impact of database domain differences between tasks, we remove domain-specific information from each sample, as detailed in Figure~\ref{fig:data_gene}. This allows us to uncover the commonalities and differences between individual tasks from a syntax perspective.
Specifically, following previous works \cite{yu2021grappa,10.1609/aaai.v37i11.26535}, we first employ the string matching method to establish entity links between questions and SQL based on the database schema $\mathcal{S}$. Subsequently, we mask all linked entities in the question to obtain $\mathcal{Q}^{de}$, and retain only the original syntactic structure in SQL to get $\mathcal{Z}$, which we named SQL skeleton. Finally, we merge these two parts to form the joint input pair $(\mathcal{Q}^{de},\mathcal{Z})$ for further analysis.

\paragraph{Component Bias Analysis} 
After the above process, we analyze component bias in the current task $\mathcal{D}^{t}$ and all historical tasks $\cup_{i=1}^{t-1}\mathcal{D}^{i}$, identifying features present in historical tasks but absent in the current one, as shown in Figure~\ref{fig:data_gene}.
Specifically, given the program-like attributes of SQL, we first utilize CodeT5 \cite{wang-etal-2021-codet5} to encode $(\mathcal{Q}^{de},\mathcal{Z})$ from all training samples of the current task to obtain representations for $K$-means clustering. Then, we extract the SQL skeleton $\mathcal{Z}$ corresponding to the sample closest to each cluster center, forming the component feature set for task $\mathcal{D}^{t}$, denoted as $\mathcal{A}^{(t)}=\{\mathcal{Z}_1^{(t)},\mathcal{Z}_2^{(t)},...,\mathcal{Z}_K^{(t)}\}$, where $K$ is the number of cluster centers. 
Finally, we select individuals that are present in the stored component feature sets of all previous tasks $\cup_{i=1}^{t-1}\mathcal{D}^{i}$ but absent in the current task $\mathcal{D}^{t}$, forming the component bias $\Delta\mathcal{A}^{(t)}$. This bias is then used to guide subsequent memory reconstruction, mitigating the model's forgetting of such features. Appendix A.1 provides more details.
When $t>1$, the formula is as follows:
\begin{equation}
\Delta\mathcal{A}^{(t)} = (\mathcal{A}^{(1)}\cup\mathcal{A}^{(2)}\cup...\cup\mathcal{A}^{(t-1)})-\mathcal{A}^{(t)}
\end{equation}
Importantly, to get $\Delta\mathcal{A}^{(t)}$, we need to save each task's $\mathcal{A}^{(t)}$, which only contains the SQL skeleton related to syntax and does not involve any real historical database information or samples, ensuring minimal storage costs. This highlights our method's feature of not requiring data replay.

\paragraph{Memory Completion} 
We employ the obtained component bias $\Delta\mathcal{A}^{(t)}$ to guide LLMs to generate pseudo-samples for the current task. This process aims to help the model fill in the memory gaps between the current and past tasks, while also uncovering knowledge relevant to the current domain from LLMs.
Unlike previous strategies that generate pseudo-data through semi-supervised learning methods such as self-training \cite{10.1609/aaai.v37i11.26492}, these methods require collecting task-related questions in advance and then constructing pseudo labels for them. In reality, obtaining accurate pseudo labels with limited annotated data for complex tasks like semantic parsing is challenging, even for most LLMs \cite{10.1145/3654930}. Therefore, we propose a soft pseudo-sample construction strategy that uses SQL skeletons to guide LLMs to simultaneously generate the natural language question $\hat{\mathcal{Q}}$ and corresponding SQL query $\hat{\mathcal{Y}}$ on the current task's database schema $\mathcal{S}^{(t)}$. Prompt details are in Appendix A.2. The generation process is as follows:
\begin{equation}
(\hat{\mathcal{Q}},\hat{\mathcal{Y}})^{(t)}= \left\{
\begin{array}{ll}
\mathcal{F}_{\text{LLM}}(\mathcal{A}^{(t)},\mathcal{S}^{(t)}), & \text{if } t = 1, \\
\mathcal{F}_{\text{LLM}}(\Delta\mathcal{A}^{(t)},\mathcal{S}^{(t)}), & \text{if } t > 1.
\end{array}
\right.
\end{equation}
Where $(\hat{\mathcal{Q}},\hat{\mathcal{Y}})^{(t)}$ refers to the pseudo samples generated by LLMs. On the first task, we use SQL skeletons $\mathcal{A}^{(t)}$ for better initialization. Each SQL skeleton generates $N_{ske}$ data.

\paragraph{Memory Calibration}
Although our proposed soft pseudo-sample construction strategy avoids the challenge of directly requiring the model to generate complex SQL queries and further leverages the generative capabilities of LLMs, it also faces the issue of hallucinations \cite{10.1145/3571730}, which can lead to noise. To address this, we design a memory calibration strategy that includes two stages: iterative self-correction and SQL skeleton-based sampling. This aims to improve the accuracy of generated pseudo-samples and ensure fidelity to specific SQL skeletons.

The inspiration for iterative self-correction comes from previous works that used LLMs for code debugging and fixing \cite{NEURIPS2023_72223cc6, chen2024teaching}. 
We iteratively execute SQL queries and correct the generated pseudo-samples, ultimately retaining only the samples that LLMs consider correct and whose SQL queries can be executed successfully. 
Additionally, we sample pseudo-samples that best match the specified SQL skeleton by calculating the edit distance after removing domain information.
Details on memory calibration are provided in Appendix A.3.

\subsubsection{Intra-Task Memory Reinforcement}
Following previous work \cite{yu2021grappa}, we introduce a data synthesis method based on context-free grammar, utilizing the component features $\mathcal{A}^{(t)}$ of the current task $\mathcal{D}^{t}$ to enhance model memory.
Specifically, for each annotated sample, we synthesize $N_{cfg}$ new instances $\hat{X}_{cfg}^{(t)}$ by synchronously replacing entities in the natural language question $\mathcal{Q}$ and SQL query $\mathcal{Y}$ with other database content based on entity linking results, as shown in Figure~\ref{fig:data_gene}. To ensure quality, only SQL queries that pass execution tests are retained. Additionally, we use LLMs to rephrase questions to enhance data diversity.

\begin{figure}[ht]
  \includegraphics[width=\columnwidth]{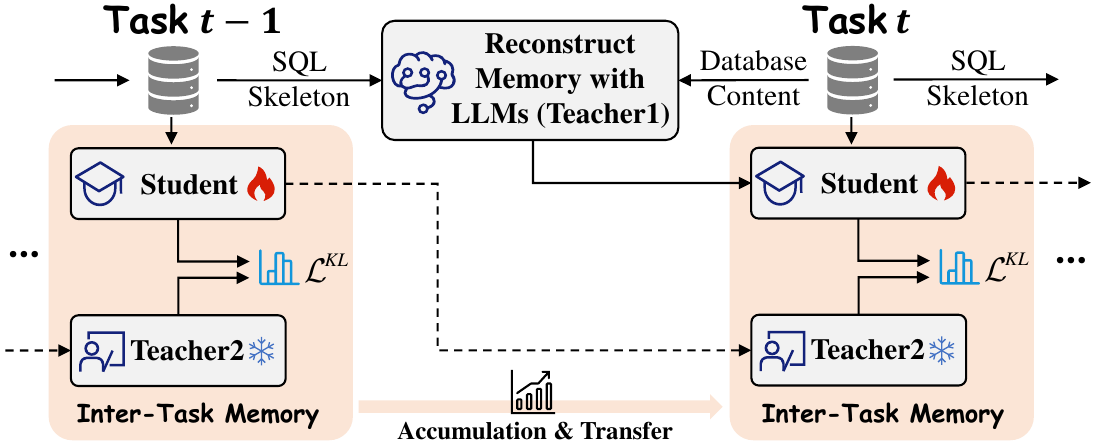}
  \caption{The task-aware dual-teacher distillation learning framework of \textsc{Lecsp}.}
  \label{fig:framework}
\end{figure}

\subsection{Task-Aware Dual-Teacher Distillation learning Framework}
\label{sec:kd}
After reconstructing the relevant inter-task and intra-task memories, we design a task-aware dual-teacher distillation framework to promote the efficient utilization of these memories. The training framework is shown in Figure~\ref{fig:framework}. Specifically, for each task, a \emph{Student} model is trained to learn from two \emph{Teacher} models: one from the LLM and the other from the previous student.

\paragraph{LLMs as Teacher 1} 
This process facilitates knowledge transfer from the LLMs (\emph{Teacher 1}) to the smaller model (\emph{Student}) via reconstructed memory, consisting of two types of pseudo data: $\hat{X}_{ske}$ and $\hat{X}_{cfg}$. They respectively represent shared knowledge accumulated from historical tasks and specific domain knowledge related to the current task. 
We utilize these pseudo datasets for additional supervised training of the smaller student model. The loss function is defined as follows:
\begin{equation}
\begin{aligned}
\mathcal{L}_{\text{cur/past}} &= \frac{1}{N^{c/s}} \sum_{i=1}^{N^{c/s}} \ell(\mathcal{F}_{\theta}(\hat{x}_i^{c/s}), \hat{y}_i^{c/s}) 
\end{aligned}
\end{equation}
Where $N^{c}, N^{s}$ respectively represent the number of data generated for types $\hat{X}_{cfg}$ and $\hat{X}_{ske}$. 
$\hat{x}_i$ and $\hat{y}_i$ with different superscripts denote the input and target output for the corresponding data types. $\ell$ means the cross-entropy loss.
 
\paragraph{Previous Student as Teacher 2} 
Although $\hat{X}_{ske}$ already covers knowledge from past tasks that is missing in the current task, this knowledge is primarily derived from indirect transfer by the teacher LLM, which may affect the stability and introduce inherent biases from the LLM. Therefore, we introduce another \emph{Teacher 2} (the student model from the previous task) to directly constrain the learning of the current student.
We encourage both models to maintain consistent representations of the shared knowledge in memory $\hat{X}_{ske}$. Specifically, inspired by previous work \cite{qin2022lfpt}, after freezing the parameters of the student model from the previous task, we use KL divergence as the distillation loss to promote consistency in the output distributions of the two models. The specific process is as follows:
\begin{equation}
    \mathcal{L}^{KL}_{past} = \frac{1}{N^{s}}\sum_{i=1}^{N^{s}} \sum_{j=1}^{l} D_\text{KL}(P_j(\hat{x}_i^{s}, \theta') \lvert \lvert P_j(\hat{x}_i^{s}, \theta))
\end{equation}
Where $l$ is the number of tokens in the output $\hat{y}_i^{s}$ corresponding to the input $\hat{x}_i^{s}$, and $P_j(\cdot)$ denotes the probability distribution over the vocabulary of model $\mathcal{F}(\cdot)$ when generating the $j$-th token. $\theta$ and $\theta'$ respectively represent the parameters of the current and previous task student models. Note that we do not calculate  $\mathcal{L}^{KL}_{past}$ on the first task.

\paragraph{Total loss function}
The loss of the student model on the original annotated data follows a similar calculation process as previously described, and it can be represented as $\mathcal{L}_{{task}} = \frac{1}{N} \sum_{i=1}^{N} \ell(\mathcal{F}_{\theta}(x_i), {y}_i)$, $N$ is the count of original annotated data. 
The final loss is as follows, where $\lambda$ represents the weight factor:
\begin{equation}
\mathcal{L}=\mathcal{L}_{{task}}+ \mathcal{L}_{cur}+\mathcal{L}_{past}+\lambda\mathcal{L}^{KL}_{past}
\end{equation}

\section{Experiments}
\paragraph{Datasets} 
We evaluate our method on the publicly available {Spider-stream-semi} \cite{10.1609/aaai.v37i11.26492} and {Combined-stream} \cite{NEURIPS2023_398b00a0} datasets.
Spider-stream-semi is derived from Spider dataset\footnote{https://yale-lily.github.io/spider} and consists of 10 tasks involving complex SQL syntax such as $\mathtt{JOIN}$ and $\mathtt{GROUP}~\mathtt{BY}$. Each task also provides unlabeled data intended for semi-supervised learning baselines, but our approach does not require these data. Most tasks have fewer than 500 annotated samples to evaluate the impact of database domain changes under limited resources.
Combined-stream consists of Spider and WikiSQL datasets\footnote{https://github.com/salesforce/WikiSQL}, with a total of 7 tasks. Unlike Spider-stream-semi, it introduces simple SQL syntax for single-table queries to assess the dual impact of table domain and SQL structure changes. 
In previous works, to ensure good initial performance, both Spider-stream-semi and Combined-stream set the task with the most annotated data as the first task, termed a \emph{warm start}. Conversely, we also introduce a variant called \emph{cold start}, which is more reflective of real-world scenarios.
In this variant, tasks are randomly shuffled without assigning the first task the most annotated data. This helps us further study the effects of initial performance and task order on the model.
Appendix B.1 provides more dataset details.

\paragraph{Evaluation Metrics}
Following prior work \cite{10.1609/aaai.v37i11.26492,NEURIPS2023_398b00a0,yu-etal-2018-spider}, we define $a_{m, n}$ as the model's accuracy on the $n$-th task's test set after training on the $m$-th task, split into Exact-set-Match (EM) and EXecution (EX) accuracy. EM assesses the formal accuracy of SQL queries and EX measures their execution results. 
(1) \emph{Average accuracy}: 
$\text{ACC}_{\text{a}}=\frac{1}{M} \sum_{i=1}^{M} a_{M,i}$. It reflects the model's overall performance across all historical tasks.
(2) \emph{Whole accuracy}: $\text{ACC}_{\text{w}}=a_{\mathcal{D}_{test}^{(1:M)}}$, where $\mathcal{D}_{test}^{(1:M)}=\cup_{i=1}^{M}\mathcal{D}^i_{test}$. It measures accuracy on all tasks' combined test sets after training on task $M$, reflecting historical performance changes.
(3) \emph{Backward transfer}: 
$\text{BWT}=\frac{1}{M-1}\sum_{i=1}^{M-1}(a_{M,i}-a_{i,i})$. It assesses the average impact of learning the $M$-th task on the performance of all previous tasks.   
(4) \emph{Forward transfer}:
$\text{FWT}=\frac{1}{M-1}\sum_{i=2}^{M}(a_{{i-1},i}-\hat{a}_i)$, where $\hat{a}_i$ represents the test accuracy of a randomly initialized reference model on the $i$-th task, examining its generalization performance.

\paragraph{Implementation Details}
In our experiments, we utilize T5-base\footnote{https://huggingface.co/google/t5-base-lm-adapt} and T5-large\footnote{https://huggingface.co/google/t5-large-lm-adapt} as backbone models, applying the Adafactor \cite{pmlr-v80-shazeer18a} optimizer. The maximum input and output lengths are configured to 512 and 256, respectively.
In the component bias analysis, we choose CodeT5\footnote{https://github.com/salesforce/CodeT5} as the encoder and set the number of cluster centers $K$ to 80. In the data generation process, we set $N_{ske}$ as 10 and $N_{cfg}$ as 3. The weight hyperparameter $\lambda$ in the training process undergoes grid search on the set \{0.03, 0.05, 0.1, 0.2, 0.3\}, and 0.1 is selected. For LLMs, We employ the publicly available Mixtral-8x7B-Instruct-v0.1, which is a sparse mixture of experts' language model. We also use other LLMs, including Meta-Llama-3-8B-Instruct and Qwen2-72B-Instruct.  
All experiments are performed on A100 GPUs with 80GB memory.

\paragraph{Baselines}
Our proposed method is compared with three types of baselines. (1) \emph{Original method}: {Sequential Fine-Tuning} (\textbf{SFT}) \cite{yogatama2019learning}. (2) \emph{Rehearsal-based methods}: 
\textbf{EMAR} \cite{han-etal-2020-continual}, \textbf{SFNET} \cite{10.1609/aaai.v37i11.26492}. (3) \emph{PET-based methods}: \textbf{PEFT} \cite{NEURIPS2023_398b00a0}, \textbf{C3} \cite{NEURIPS2023_398b00a0}, \textbf{ProgPrompt} \cite{razdaibiedina2023progressive}. 
Since our method uses LLMs as \emph{Teacher 1} for zero-shot memory construction, we also report their zero-shot performance.
The \textbf{ORACLE} setting trains the model incrementally on all tasks' data, representing the performance upper bound for CSP. 
More details on the baseline are provided in Appendix B.2.

\begin{table*}[ht]
\centering
\resizebox{\textwidth}{!}{%
\begin{tabular}{lccccclccccl}   %
\toprule

\multirow{2}{*}{\textbf{Method}} & \multicolumn{1}{c}{\multirow{2}{*}{\textbf{DR}}} & \multicolumn{5}{c}{\textbf{Spider-stream-semi}} & \multicolumn{5}{c}{\textbf{Combined-stream}} \\
\cmidrule(lr){3-12} 
 &  & {$\text{ACC}_\text{a}$} & {$\text{ACC}_\text{w}$} & {BWT} & {FWT} & {$\Delta$} & {$\text{ACC}_\text{a}$} & {$\text{ACC}_\text{w}$} & {BWT} & {FWT} & {$\Delta$}  \\ 
 \hline
\rowcolor{gray!30} \multicolumn{12}{c}{\textbf{\emph{T5-Base(220M)}}} \\
SFT & {\ding{55}} & 41.7/45.3  & 40.1/43.7  & -13.6/-11.7  & 22.5/25.2 & {\color{darkred} $\downarrow$1.5}  & 51.0/36.0  & 52.4/38.3  & -11.1/-17.6  & 19.9/11.1 & {\color{darkred} $\downarrow$11.1} \\
ProgPrompt & {\ding{55}} & 14.9/16.5 & 14.3/16.4 & -25.6/-26.3 & 14.7/15.8 & {\color{darkred} $\downarrow$14.9} & 21.3/19.2 & 27.2/26.8 & -43.9/-41.2 & 6.1/3.3 & {\color{darkred} $\downarrow$21.1} \\
EMAR & {\ding{51}} & 44.1/48.0 & 43.0/47.0 & -11.8/-9.7 & 23.7/25.2 & {\color{darkgreen} $\uparrow$2.3} &  60.6/\underline{58.9} & \underline{55.7}/\underline{55.3} & -7.9/-7.1 & \underline{28.7}/25.4 & {\color{darkred} $\downarrow$5.6} \\
SFNET$^{\clubsuit}$  & {\ding{55}} & 39.5/43.9 & 38.1/42.7 & -14.6/-12.4 & 22.7/25.7 & {\color{darkgreen} $\uparrow$0.4} & 56.2/49.4 & 51.8/48.4 & -7.9/-14.4 & 21.0/14.6 & {\color{darkred} $\downarrow$21.7} \\
SFNET$^{\clubsuit}$ & {\ding{51}} & \underline{45.8}/\underline{48.1} & \underline{44.2}/\underline{47.7} & \underline{-6.4}/\underline{-6.4} & \underline{23.9}/\underline{26.2} & {\color{darkred} $\downarrow$3.4} & \underline{60.7}/57.7 & {55.2}/53.9 & \underline{-6.0}/\underline{-6.7} & \underline{28.7}/\underline{26.0} & {\color{darkred} $\downarrow$6.7} \\
\textsc{Lecsp}(Ours) & {\ding{55}} & \textbf{47.4}/\textbf{53.7} & \textbf{46.5}/\textbf{53.3} & \textbf{-4.7}/\textbf{-4.6} & \textbf{30.1}/\textbf{32.8} & {\color{darkgreen} $\uparrow$1.4} & \textbf{63.4}/\textbf{61.7} & \textbf{60.5}/\textbf{59.8} & \textbf{-4.6}/\textbf{4.3} & \textbf{32.7}/\textbf{31.5} & {\color{darkred} $\downarrow$4.8} \\
\hline
\multicolumn{12}{c}{\emph{Ideal Setting \& Upper Bound}} \\
PEFT$^{\diamondsuit}$ & {\ding{55}} & 40.6/43.8 & {44.6}/{48.4} & 0.0/0.0 & -/- & {\color{darkred} $\downarrow$32.5} & 63.8/- & 66.2/- & 0.0/0.0 & -/- & {\color{darkred} $\downarrow$38.8} \\
ORACLE & {} & 60.3/61.3 & 62.8/63.9 & 4.6/3.3 & 25.7/27.9 & {\color{darkred} $\downarrow$1.1} & 70.2/68.2 & 71.1/70.8 & 5.6/4.7 & 29.2/26.6 & {\color{darkgreen} $\uparrow$0.3} \\
\hline\hline
\rowcolor{gray!30} \multicolumn{12}{c}{\textbf{\emph{T5-Large(770M)}}} \\
SFT & {\ding{55}} & {46.4}/{50.4} & 44.3/48.9 & -13.2/-11.5 & 29.0/31.5 & {\color{darkgreen} $\uparrow$1.8} & 57.8/48.9 & 58.9/53.3 & -7.6/-16.4 & 28.3/25.4 & {\color{darkred} $\downarrow$8.3} \\
ProgPrompt & {\ding{55}} & 22.1/23.9 & 22.4/24.6 & -27.0/-27.5 & 21.1/23.1 & {\color{darkred} $\downarrow$20.8} & 23.5/20.2 & 29.8/27.7 & -46.4/-42.7 & 10.2/3.7 & {\color{darkred} $\downarrow$22.4} \\
EMAR & {\ding{51}} & 48.9/51.8 & 48.4/52.5 & -8.5/-8.3 & 30.3/33.5 & {\color{darkgreen} $\uparrow$1.4} & 63.9/61.9 & 60.6/60.4 & -6.6/-8.9 & 33.8/29.3 & {\color{darkred} $\downarrow$5.5} \\
SFNET$^{\clubsuit}$ & {\ding{55}} & 44.2/48.2 & 43.3/47.8 & -12.3/-9.9 & 29.9/31.0 & {\color{darkgreen} $\uparrow$3.3} & 65.0/60.3 & 61.5/59.7 & -7.5/-11.9 & 28.2/20.4 & {\color{darkred} $\downarrow$24.1} \\
SFNET$^{\clubsuit}$ & {\ding{51}} & \underline{52.2}/\underline{55.0} & \underline{52.0}/\underline{55.8} & \underline{-6.5}/\textbf{-4.2} & \underline{34.1}/\underline{35.5} & {\color{darkred} $\downarrow$2.2} & \underline{65.3}/\underline{64.7} & \underline{61.8}/\underline{61.9} & \underline{-5.7}/\underline{-5.2} & \underline{34.9}/\underline{31.1} & {\color{darkred} $\downarrow$5.6} \\ 
\textsc{Lecsp}(Ours) & {\ding{55}} & \textbf{58.6}/\textbf{63.5} & \textbf{57.4}/\textbf{63.8} & \textbf{-5.8}/\underline{-5.0} & \textbf{41.4}/\textbf{44.3} & {\color{darkgreen} $\uparrow$0.8} & \textbf{68.2}/\textbf{66.7} & \textbf{66.5}/\textbf{65.4} & \textbf{-2.9}/\textbf{-3.7} & \textbf{37.1}/\textbf{34.5} & {\color{darkred} $\downarrow$5.1} \\
\hline
\multicolumn{12}{c}{{\emph{Ideal Setting \& LLMs Performance \& Upper Bound}}} \\
PEFT$^{\diamondsuit}$ & {\ding{55}} & 49.6/52.4 & {53.4}/{56.4} & 0.0/0.0 & -/- & {\color{darkred} $\downarrow$24.1} & 67.3/- & 70.0/- & 0.0/0.0 & -/- & {\color{darkred} $\downarrow$23.4} \\
C3$^{\diamondsuit}$ & {\ding{55}} & -/- & -/- & -/- & -/- & {-} & 69.0/- & 71.2/- & 0.0/0.0 & -/- & {-} \\
C3$^{\diamondsuit\dagger}$ & {\ding{55}} & -/- & -/- & -/- & -/- & {-} & 67.6/- & 70.0/- & 0.0/0.0 & -/- & {-} \\
Mixtral-8x7B & {\ding{55}} & 22.0/49.4 & 22.2/52.4 & -/- & -/-  & {-} & 28.6/24.4 & 27.0/26.0 & -/- & -/- & {-} \\
ORACLE & {} & 66.6/68.2 & 68.6/70.1 & 3.7/4.0 & 34.3/36.0 & {\color{darkgreen} $\uparrow$0.4} & 73.7/73.2 & 75.8/76.0 & 1.9/2.3 & 37.3/34.2 & {\color{darkred} $\downarrow$0.1}  \\
\bottomrule
\end{tabular}
}
\caption{Results (EM/EX) on Spider-stream-semi and Combined-stream datasets (\%). \textsc{Lecsp} uses Mixtral-8x7B as \emph{Teacher 1}. DR indicates whether historical data replay is used, with a memory size set to 10.
$\Delta$ represents the performance change ({$\text{ACC}_\text{a}$}--EM) from the current \emph{warm start} to the \emph{cold start} scenario, detailed results can be found in Appendix C.1. 
$\clubsuit$ indicates the need for additional unsupervised data, and ${\diamondsuit}$ means an ideal continuous learning setting is used, but forward transfer is impossible. The results of PEFT and C3 on the Combined-stream are from the original paper and its official code repository. $\dagger$ signifies the use of GPT-3.5 as the teacher. The best results are highlighted in bold, and the second-best results are underlined. Our results are the average of three random runs.}
\label{tab:main_result}
\end{table*}

\section{Results and Analysis}
\subsection{Overall Results}
Table \ref{tab:main_result} reports the overall performance of each method on two benchmarks under different backbones. More results and analysis are provided in Appendix C.
\paragraph{Comparison with PET-based Methods}
\textsc{Lecsp} outperforms PEFT by up to 11.1\%. PEFT uses an ideal continual learning setup that sacrifices model generalization, as it is challenging to select the appropriate PET module for unseen samples, making FWT calculation impossible. 
We also note that ProgPrompt does not require ideal settings, but its performs worse than SFT. This is because ProgPrompt is designed for tasks with simpler output structures, such as classification. However, for models pre-trained on natural language corpora, adapting to complex tasks like semantic parsing with limited-length prompts is challenging \cite{qin2022lfpt}. The progressive prompt design of ProgPrompt further compresses the prompt length for individual tasks, worsening the issue and making it nearly non-functional in \emph{cold start} scenarios.
Notably, PET-based methods with ideal settings, like PEFT, heavily rely on the initial task's labeled data. In \emph{cold start} scenarios, their performance drops sharply, with a maximum decline of 38.8\%.

\paragraph{Comparison with Rehearsal-based Methods}
Compared to the previous SOTA method SFNET, \textsc{Lecsp} achieves an improvement of up to 8.8\%, despite SFNET using additional unlabeled data.
${\text{SFNET}}$ (\emph{w/o} DR) performs worse than SFT, which may be attributed to the intensified impact of noisy pseudo labels in self-training, indicating SFNET's reliance on data replay. In contrast, our approach excels without data replay. Compared to EMAR, \textsc{Lecsp} also achieves a maximum gain of 11.7\%.

\paragraph{Comparison with ORACLE and LLMs}
Impressively, our method exceeds most performance upper bound on FWT with up to 8.3\%, demonstrating effective knowledge transfer from LLMs to smaller models and enhancing generalization capabilities. In contrast, LLMs alone perform worse. 
We also find that different types and scales of LLMs effectively enhance student models, with more results in Appendix C.2.

\subsection{Detailed Results and Analysis}

\begin{figure*}[ht]
\centering
\includegraphics[width=\textwidth]{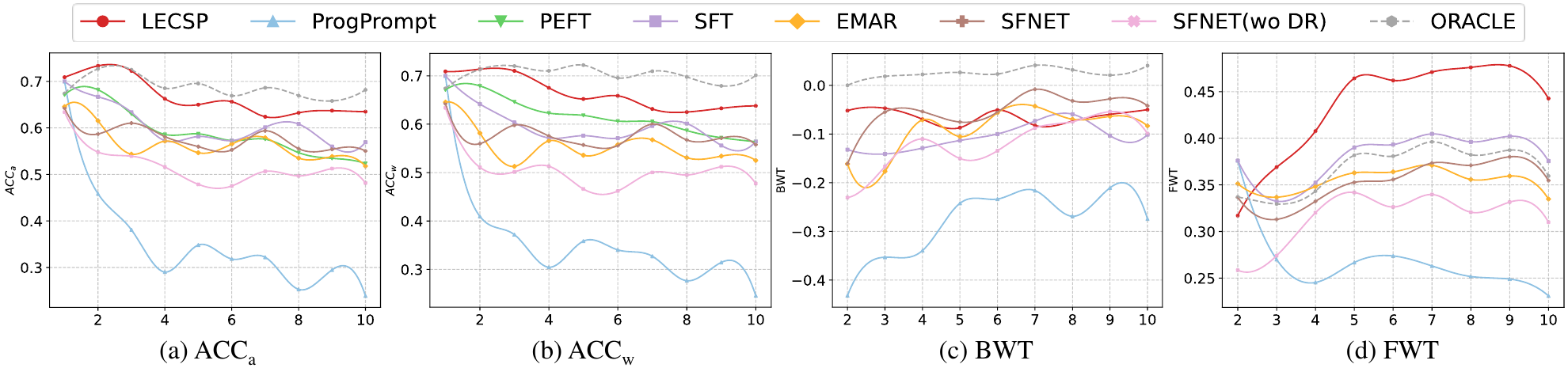}
\caption{Results (EX) till the seen tasks based on Spider-stream-semi dataset (T5-large).
} 
\label{fig:result_line}
\end{figure*}

\paragraph{Results Till the Seen Tasks}
Figure~\ref{fig:result_line} shows the performance of different methods on previously seen tasks. As task numbers increase, \textsc{Lecsp} (red) consistently remains closest to the upper bound ORACLE (grey) on $\text{ACC}_{\text{a}}$ and $\text{ACC}_{\text{w}}$. It outperforms most baselines in BWT with stable performance, while maintaining a significant advantage in FWT.
Appendix C.3-C.4 provides more detailed results.

\paragraph{Influence of Task Order}
Unlike previous works that start with data-rich tasks, we further explore data-scarce cold-start scenarios. The results in Appendix C.1 indicate a significant performance drop for most methods, especially for \emph{PET-based} methods, reflecting their dependence on the performance of the first task. In contrast, \textsc{Lecsp} demonstrates robustness and achieves SOTA performance, outperforming SFNET with data replay by up to 11.4\%.

\begin{figure}[ht]
  \includegraphics[width=\columnwidth]{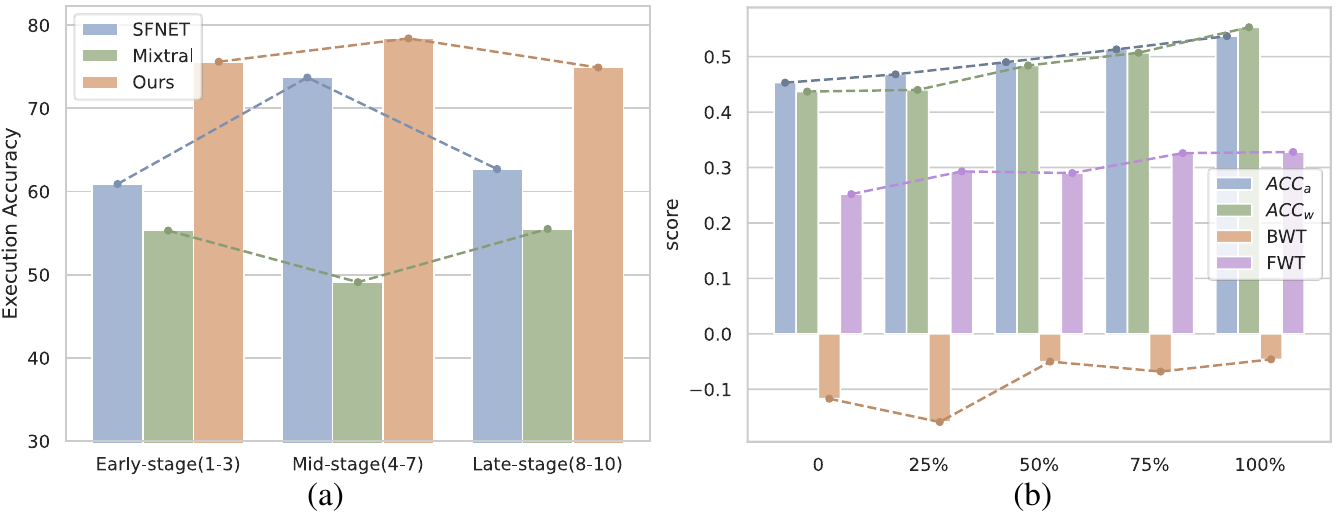}
  \caption{(a) Accuracy of synthetic data execution across different stages. (b) Impact of pseudo sample quantity on different metrics (T5-base).}
  \label{fig:quality}
\end{figure}

\begin{table}[ht]
\centering
\resizebox{0.48\textwidth}{!}{%
\begin{tabular}{lcccc}
\toprule
{} & \multicolumn{1}{c}{$\text{ACC}_{\text{a}}$} & \multicolumn{1}{c}{$\text{ACC}_{\text{w}}$} & \multicolumn{1}{c}{BWT} & \multicolumn{1}{c}{FWT} \\ 
\hline
$\text{S}_{\text{S}}$ & {47.4}/{53.7} & {46.5}/{53.3} & {-4.7}/{-4.6} & {30.1}/{32.8} \\
$\text{S}_{\text{C}}$ & {45.0}/{49.1} & {43.7}/{48.0} & {-4.5}/{-5.7} & {28.1}/{31.7} \\
\hline
$\text{C}_{\text{C}}$ & {48.8}/{52.6} & {49.9}/{54.2} & {-1.1}/{-0.9} & {24.8}/{27.3} \\ 
$\text{C}_{\text{S}}$ & {43.6}/{48.1} & {44.2}/{49.2} & {-4.4}/{-5.6} & {27.1}/{30.2} \\
\bottomrule
\end{tabular}
}
\caption{The impact of SQL syntax variance on model performance (\%).
$\text{S}_{\text{S}}$ denotes using component bias from \underline{\textbf{S}}pider-stream-semi itself, as does $\text{C}_{\text{C}}$. 
$\text{S}_{\text{C}}$ indicates using component bias from Spider-stream-semi (\underline{\textbf{C}}old start) to construct memory for \underline{\textbf{S}}pider-stream-semi, and vice versa.} 
\label{tab:sql_variance}
\end{table}

\paragraph{Influence of SQL Syntax Variance}
To further explore the role of SQL syntax variance in \textsc{Lecsp}, we compare the impact on model performance between our method and randomly obtained component bias. Specifically, we swap the memories constructed under different orders of Spider-stream-semi and Spider-stream-semi (\emph{cold start}), keep other experimental settings unchanged, and retrain the models. The results in Table \ref{tab:sql_variance} show that, after the exchange, model performance decreases on most metrics but improves in FWT on $\text{C}_{\text{S}}$. This improvement may be due to the introduction of unseen SQL syntax, enhancing generalization.

\paragraph{Quality and Quantity of Pseudo Samples}
To evaluate the quality of generated memories (pseudo samples), we compare our method with the self-training method used in SFNET (T5-large) and the zero-shot results of Mixtral, both using additional unsupervised data. We divide the original Spider-stream-semi into early, mid, and late stages, randomly selecting 70 pseudo-samples from each stage to compare SQL execution accuracy. For our method, we use manual evaluation, whereas for the other two methods, we rely on their original labels. The results in Figure~\ref{fig:quality}(a) show that our method significantly improves the quality of pseudo-samples, while the self-training method in SFNET is less stable. Figure~\ref{fig:quality}(b) shows that as the proportion of pseudo-samples increases, all metrics generally improve, but several metrics stabilize in the later stages. 
Details on the number of pseudo-samples constructed for intra-task and inter-task are provided in Appendix B.1.

\begin{table}[ht]
\centering
\resizebox{0.48\textwidth}{!}{%
\begin{tabular}{lcccc}
\toprule
{} & \multicolumn{1}{c}{$\text{ACC}_{\text{a}}$} & \multicolumn{1}{c}{$\text{ACC}_{\text{w}}$} & \multicolumn{1}{c}{BWT} & \multicolumn{1}{c}{FWT} \\ 
\hline
{\textsc{Lecsp}}  & 47.4/53.7  & 46.5/53.3  & -4.7/-4.6  & 30.1/32.8 \\
\hline
{ - $\hat{X}_{cfg}$} &{45.6}/{50.3} & {44.3}/{49.1} & {-5.6}/{-6.6} & {27.7}/{30.7} \\
{ - MC} & {20.3}/{22.9} & {19.4}/{21.9} & {0.4}/{-0.6} & {15.8}/{17.5} \\ 
{ - {T2}} & {46.6}/{52.8} & {45.5}/{51.8} & {-6.6}/{-5.9} & {30.4}/{33.3} \\
{ - $\hat{X}_{ske}$} & 41.7/45.3 & 40.1/43.7 & -13.6/-11.7 & 22.5/25.2 \\
\bottomrule
\end{tabular}
}
\caption{Ablation study results on Spider-stream-semi.}
\label{tab:ablation}
\end{table}

\paragraph{Ablation Study}
We decompose \textsc{Lecsp} (T5-base) and sequentially remove each component to evaluate its impact. 
Table \ref{tab:ablation} shows that \textsc{Lecsp} with all modules delivers the best overall performance. 
Omitting intra-task memory $\hat{X}_{cfg}$ or inter-task memory $\hat{X}_{ske}$ reduces performance. Particularly for $\hat{X}_{ske}$, which includes relevant past knowledge, bringing the most significant gains (5.7\%-9.6\%).
Abandoning Memory Calibration (MC) strategy causes a steep performance drop, up to 31.4\%, indicating that MC effectively mitigates hallucination \cite{10.1145/3571730} issues in LLM-generated memories.
Ignoring \emph{Teacher 2} (T2) leads to an overall performance decline (0.8\%-1.9\%) despite minor gains on FWT (0.3\%-0.5\%). 
Appendix C.5 provides detailed results on the hyperparameters, including the number of clusters $K$ and the weighting factor $\lambda$.

\section{Conclusion}
In this paper, we introduce \textsc{Lecsp}, a novel CSP method without real data replay. 
\textsc{Lecsp} analyzes task commonalities and differences to extract key SQL syntax features, guiding LLMs to generate pseudo samples to reconstruct memory. we also design a task-aware dual-teacher distillation framework to facilitate the transfer of LLMs' internal knowledge and historical knowledge to small models. Extensive experimental results demonstrate that \textsc{Lecsp} significantly outperforms baselines in various task scenarios, even if they use data replay or ideal settings. Furthermore, \textsc{Lecsp} surpasses the upper bound and demonstrates adaptability to challenging real-world scenarios.

\bibliography{anonymous-submission-latex-2025}

\section{A\quad Method Details}
\subsection{A.1\quad Component Bias Analysis}
First, we utilize CodeT5 \cite{wang-etal-2021-codet5}, which is pre-trained on programs, to acquire the high-dimensional representations $\mathbf{h}$ for all joint inputs $(\mathcal{Q}^{de},\mathcal{Z})$.
Second, we perform $K$-means clustering on all training samples of the current task $\mathcal{D}^i$ and obtain the cluster centers $\mathbf{O}^{(i)}$.
\begin{equation}
\mathbf{O}^{(i)} = \text{K-Means}(\mathbf{H}^{(i)}, K)
\end{equation}
Where $\mathbf{O}^{(i)}=\{\mathbf{o}_1^{(i)},\mathbf{o}_2^{(i)},...,\mathbf{o}_K^{(i)}\}$ is the set of cluster centers, and the set of representations for all individuals in $\mathcal{D}^i$ is denoted as $\mathbf{H}^{(i)}=\{\mathbf{h}_1^{(i)},\mathbf{h}_2^{(i)},...,\mathbf{h}_{|\mathcal{D}^i|}^{(i)}\}$. $K$ is the predefined number of clusters.

Then, for each task $\mathcal{D}^i$ and its set of cluster centers $\mathbf{O}^{(i)}$, we identify the individual closest to each $\mathbf{o}^{(i)}_j$ and obtain its SQL skeletons $\mathcal{Z}_j^{(i)}$ through the mapping function $f(\cdot)$, resulting in a task feature set $\mathcal{A}^{(i)}=\{\mathcal{Z}_1^{(i)},\mathcal{Z}_2^{(i)},...,\mathcal{Z}_K^{(i)}\}$.  
\begin{equation}
\mathcal{Z}_j^{(i)} = f(\arg\min_{\mathbf{h} \in \mathbf{H}^{(i)}} \lVert \mathbf{h} - \mathbf{o}^{(i)}_j \rVert), \quad \forall \mathbf{o}^{(i)}_j \in \mathbf{O}^{(i)}
\end{equation}

\begin{figure*}[ht]
\centering
\includegraphics[width=0.95\textwidth]{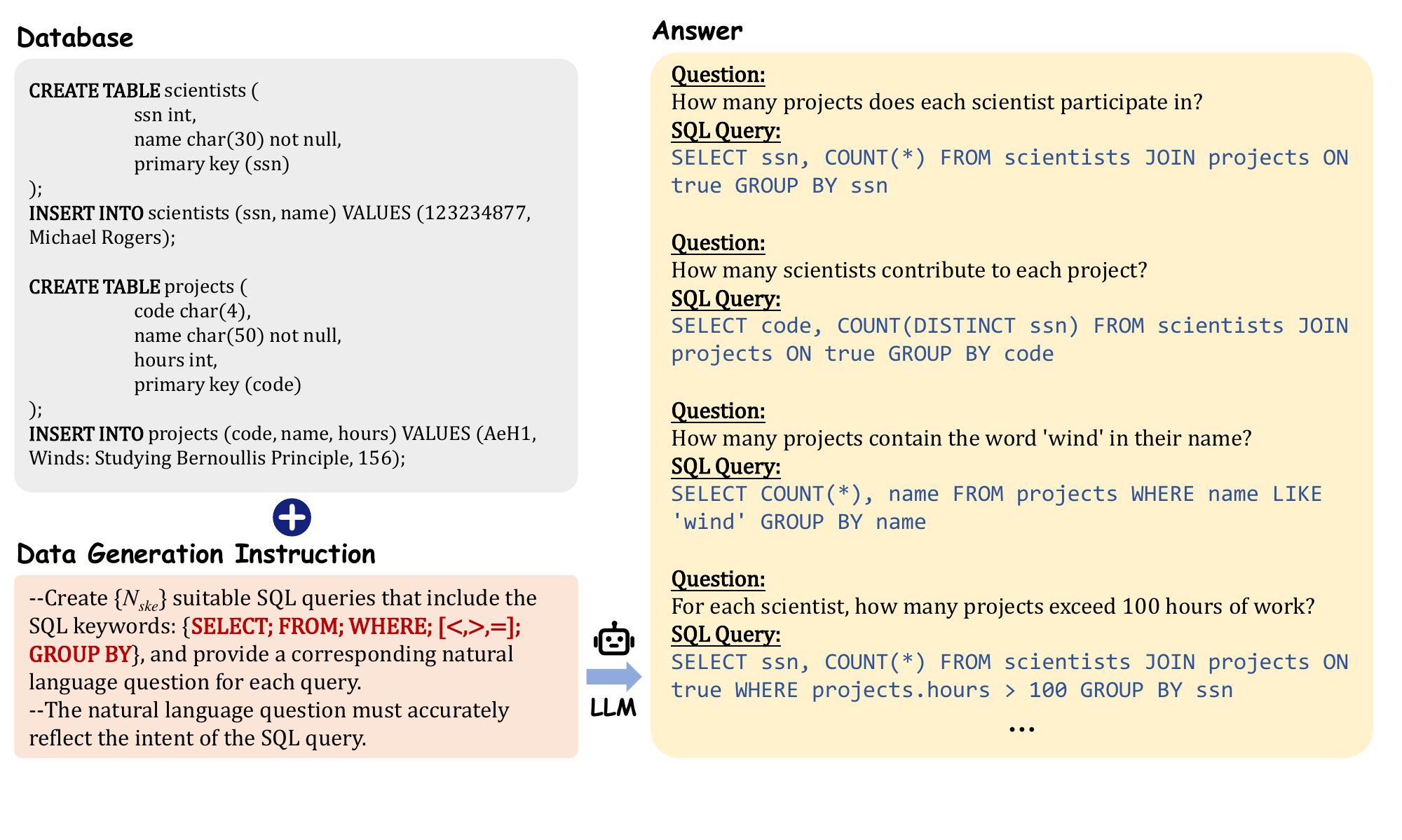}
\caption{The prompt case of inter-task memory completion based on SQL skeleton.} 
\label{fig:prompt_gene}
\end{figure*}

\begin{figure*}[ht]
\centering
\includegraphics[width=0.95\textwidth]{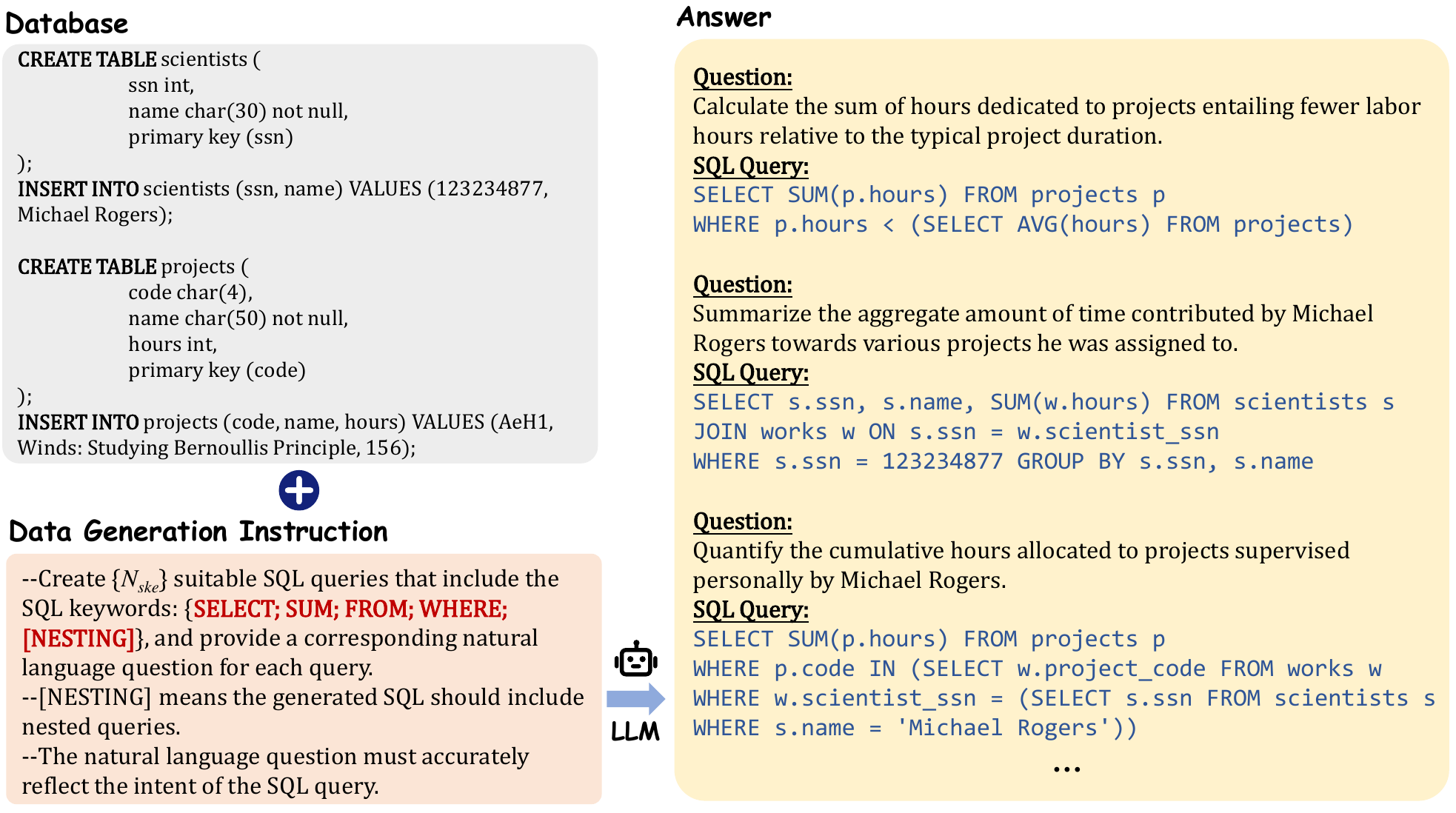}
\caption{The prompt case of inter-task memory completion based on SQL skeleton (with nested structures).} 
\label{fig:prompt_gene_nest}
\end{figure*}

\subsection{A.2\quad Inter-Task Memory Completion}
\label{apx:inter_ask_memory}
We follow the database schema style suggested in previous work \cite{chang2023how} and randomly select one row of data from each database table. 
we observe that generating question-SQL pairs strictly adhering to the SQL skeleton poses challenges for LLMs. 
To reduce complexity, as shown in Figure~\ref{fig:prompt_gene}, we simplify the original SQL skeleton by retaining only the unique SQL keywords such as \{$\mathtt{SELECT}$ [$\mathtt{COL1}$] $\mathtt{FROM}$ [$\mathtt{TAB1}$] $\mathtt{WHERE}$ [$\mathtt{COL2}$] = [$\mathtt{VAL1}$] $\mathtt{GROUP~BY}$ [$\mathtt{COL1}$]\} $\rightarrow$ \{$\mathtt{SELECT;FROM;WHERE;}$[$\mathtt{<,>,=}$]$\mathtt{;GROUP~BY}$\}. 
If the original SQL skeleton contains nested syntax, we provide nested structure hints, such as \{$\mathtt{NESTING}$\}, as illustrated in Figure~\ref{fig:prompt_gene_nest}. 
Note that during the sampling stage of memory calibration, we still use the complete SQL skeleton as a reference to choose the structurally closest pseudo-samples.

\begin{algorithm}[ht]
\caption{Iterative Self-Correction}
\label{alg:Self-Correction}
\begin{algorithmic}[1]
\renewcommand{\algorithmicrequire}{\textbf{Input:}}
\renewcommand{\algorithmicensure}{\textbf{Output:}}
\REQUIRE \raggedright Natural language questions $\hat{\mathcal{Q}}$, SQL queries $\hat{\mathcal{Y}}$, database schema ${\mathcal{S}}$. 
\ENSURE \raggedright Candidate pseudo samples $\hat{X}_{ske}^{cand}$.

\STATE $\hat{X}_{input} \gets \{\hat{\mathcal{Q}}, \hat{\mathcal{Y}}, {\mathcal{S}}\}$
\STATE $\hat{X}_{ske}^{cand} \gets \{\}$

\FOR{$(\hat{q},\hat{y},s) \in \hat{X}_{input}$}
    \STATE $r \gets \textsc{Executor}(\hat{y})$ 
    \IF{ no \emph{Execution Error} in $r$ \textbf{and} $\textsc{Llm}_{verify}(\hat{q},\hat{y},s, r) = \emph{Correct}$ }  
        \STATE $\hat{X}_{ske}^{cand} \gets \hat{X}_{ske}^{cand} \cup \{(\hat{q}, \hat{y}, s)\}$
        \STATE \textbf{continue}
    \ELSE
        \FOR{$i \gets 1$ to $M$}
            \STATE $\hat{y}_{i} \gets \textsc{Llm}_{revise}(\hat{q},\hat{y},s, r) $ 
            \STATE $r_{i} \gets \textsc{Executor}(\hat{y}_{i})$
                \IF{no \emph{Execution Error} in $r_{i}$ \textbf{and} $\textsc{Llm}_{verify}(\hat{q},\hat{y}_{i},s, r_{i}) = \emph{Correct}$}
                \STATE $\hat{X}_{ske}^{cand} \gets \hat{X}_{ske}^{cand} \cup \{(\hat{q}, \hat{y}_{i}, s)\}$
                \STATE \textbf{break}
                \ENDIF
        \ENDFOR
    \ENDIF
\ENDFOR
\RETURN $\hat{X}_{ske}^{cand}$
\end{algorithmic}
\end{algorithm}

\begin{figure*}[ht]
\centering
\includegraphics[width=0.95\textwidth]{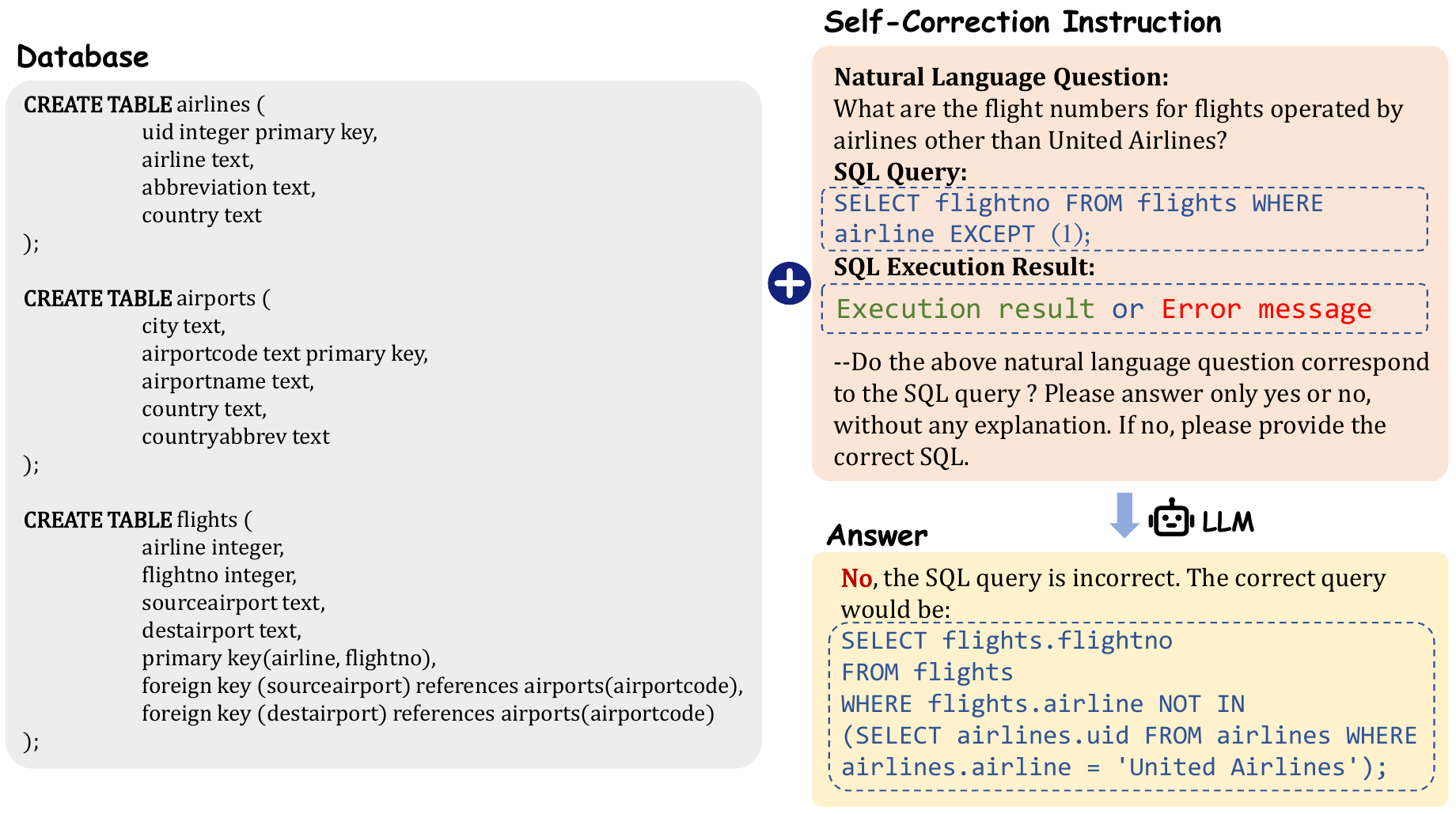}
\caption{The prompt case of self-correction. \emph{Execution result} and \emph{Error message} are returned by the SQL executor.} 
\label{fig:prompt_self_correct}
\end{figure*}

\subsection{A.3\quad Memory Calibration}
\label{apx:memory_calibration}
Memory calibration consists of two main stages: iterative self-correction and SQL skeleton-based sampling.
Iterative self-correction aims to improve the accuracy of pseudo samples, including the consistency between the generated questions and SQL, and the executability of the SQL queries, as shown in Algorithm~\ref{alg:Self-Correction}. First, for a natural language question ${\hat{q}}$ and SQL query ${\hat{y}}$, we execute ${\hat{y}}$ on the database with schema ${s}$, and get the execution result ${r}$ (line 1--line 4). Note that this result may be a correct output or an error message from the database management system. Second, we input the (${\hat{q}}$, ${\hat{y}}$, ${s}$, ${r}$) back into the LLM for verification, retaining the samples that the LLM considers correct and the SQL execution is successful (line 5--line 7). For incorrect samples, we require the LLM to revise the SQL query and set the upper limit of this process to $M$ times (line 9--line 14). In our experiment, $M$ is set to 3. The single-round self-correction prompt is shown in Figure~\ref{fig:prompt_self_correct}.

\begin{algorithm}[ht]
\caption{Sampling Based on SQL Skeleton}
\label{alg:sampling}
\begin{algorithmic}[1]
\renewcommand{\algorithmicrequire}{\textbf{Input:}}
\renewcommand{\algorithmicensure}{\textbf{Output:}}
\REQUIRE \raggedright SQL Skeleton $\mathcal{Z}$, list of candidate pseudo samples $\hat{X}_{ske}^{cand} = \{\hat{x}_1, \hat{x}_2,..., \hat{x}_{n}\}$, where $\hat{x}_{i}=(\hat{q}_{i},\hat{y}_{i},s_{i})$.
\ENSURE \raggedright Pseudo samples $\hat{X}_{{ske}}$.

\STATE $\hat{X}_{{ske}} \gets \{\}$
\STATE $ \mathcal{D} \gets \{\}$
\FOR{$\hat{x}_{i} \in \hat{X}_{ske}^{cand}$}
    \STATE $\hat{q}_{i},\hat{y}_{i},s_{i} \gets \hat{x}_{i} $
    \STATE $\hat{y}^{'}_{i} \gets \textsc{De-domian}(\hat{y}_{i})$
    \STATE $d_{i} \gets \textsc{Similarity}(\mathcal{Z}, \hat{y}^{'}_{i})$  
    \STATE $\mathcal{D} \gets \mathcal{D} \cup \{(\hat{x}_{i}, d_{i})\}$
\ENDFOR
\STATE $ \mathcal{D}^{'} \gets \textsc{Sort}(\mathcal{D})$
\STATE $\hat{X}_{{ske}} \gets \{\hat{x}|(\hat{x}, d) \in \mathcal{D}^{'}[1:R]\}  $

\RETURN $\hat{X}_{{ske}}$
\end{algorithmic}
\end{algorithm}

We also design a simple but effective sampling strategy to select the samples that best match the specified SQL skeleton from the candidate pseudo samples, as shown in Algorithm~\ref{alg:sampling}. Specifically, for each pseudo-sample's SQL query $\hat{y}$, we first remove its database-specific information according to the method described in Section: Reconstruct Memory with LLMs (line 5). Then, we calculate the edit distance between it and the given SQL skeleton $\mathcal{Z}$ (line 6 -- line 7). Finally, we select the top $R$ samples with the highest similarity (line 9 -- line 10), and $R$ is set to 3.

\section{B\quad Implementation Details}
\label{sec:Detailed_Experimental_Setup}

\subsection{B.1\quad Dataset}
\label{apx:dataset_detail}

\begin{figure*}[ht]
\centering
\includegraphics[width=0.95\textwidth]{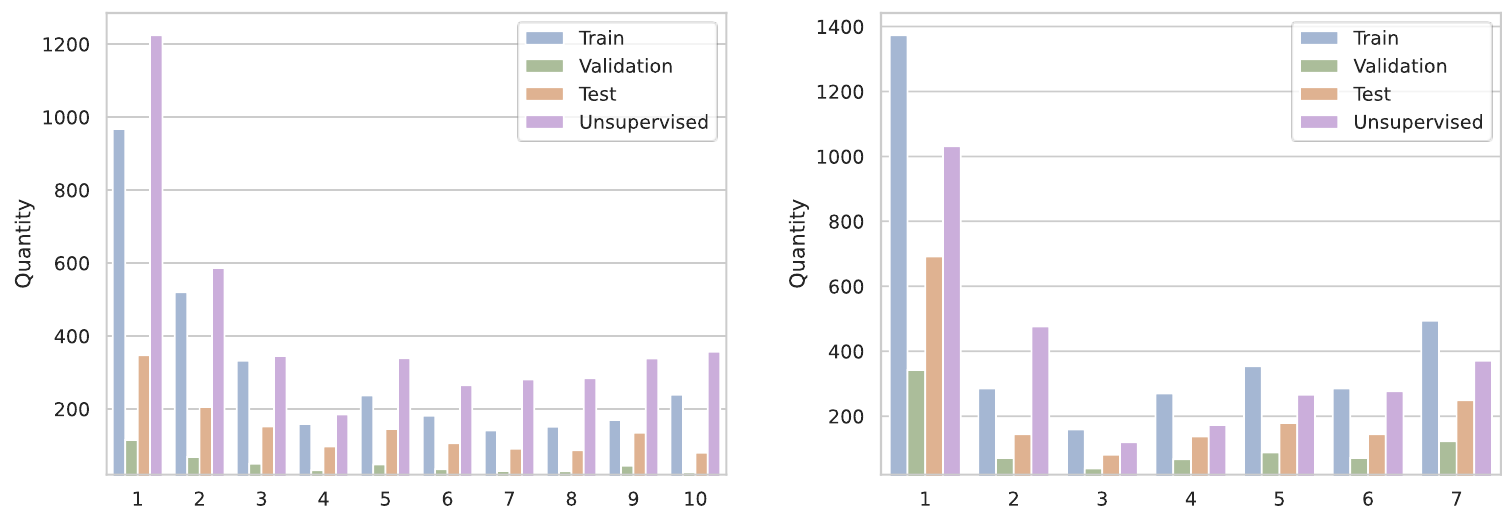}
\caption{Statistics of Spider-stream-semi (left) and Combined-stream (right).} 
\label{fig:stastic}
\end{figure*}

\begin{figure*}[ht]
\centering
\includegraphics[width=0.95\textwidth]{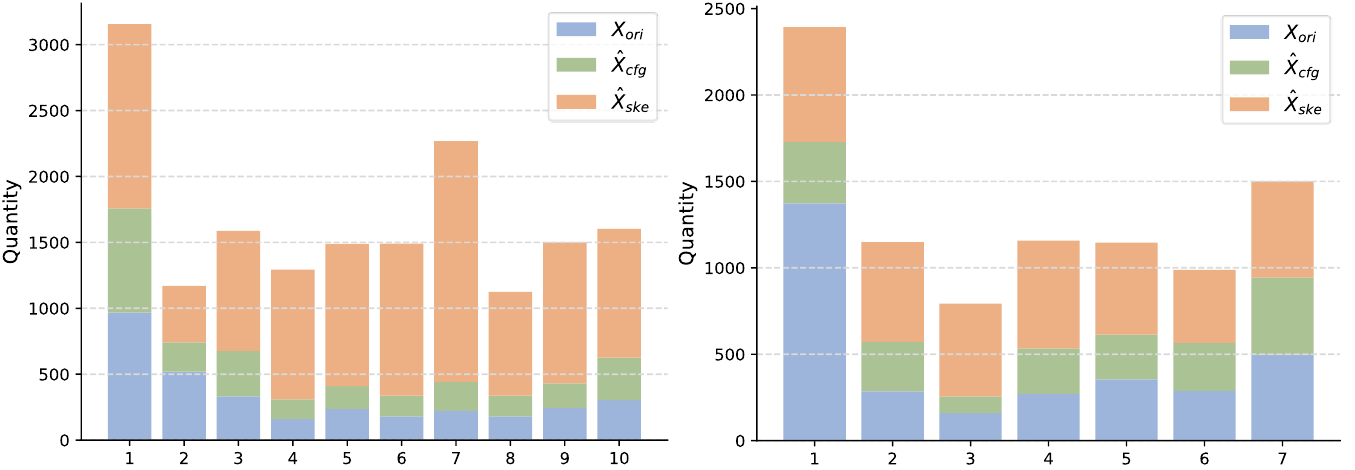}
\caption{Statistics of pseudo-samples constructed by \textsc{Lecsp} on Spider-stream-semi (left) and Combined-stream (right).} 
\label{fig:count_psedu}
\end{figure*}

The data statistics for Spider-stream-semi and Combined-stream are shown in Figure~\ref{fig:stastic}. The order of Spider-stream-semi (\emph{cold start}) compared to its original version is [3, 5, 7, 0, 2, 6, 8, 4, 9, 1], and the order of Combined-stream (\emph{cold start}) compared to its original version is [3, 2, 4, 0, 6, 5, 1]. 
We note that both \citet{10.1609/aaai.v37i11.26492} and \citet{NEURIPS2023_398b00a0} divide streaming benchmarks based on the Spider dataset. They share similar features and objectives, with the main difference being that the former provides additional unlabeled data to support semi-supervised methods like SFNET. Our approach does not require this data. To make the above baselines work well, we use the benchmark of \citet{10.1609/aaai.v37i11.26492} and name it Spider-stream-semi.
Since the original Combined-stream dataset does not provide unsupervised data, we follow the work of \citet{NEURIPS2023_398b00a0}, using unused data from Spider and WikiSQL as additional unsupervised data to ensure SFNET functions properly.
The statistics of the number of pseudo-samples in our method are shown in Figure~\ref{fig:count_psedu}.

\subsection{B.2\quad Baselines}
\label{apx:baseline_detail}
Consistent with most previous semantic parsing work \cite{xie-etal-2022-unifiedskg, 10.1609/aaai.v37i11.26535, Li_Hui_Cheng_Qin_Ma_Huo_Huang_Du_Si_Li_2023}, we uniformly adopt the T5 series models as the backbone and use the comprehensive evaluation metrics of Exact-set-Match (EM) and Execution accuracy (EX).
In our method, the version based on T5-base uses a batch size of 14 with 40 epochs, while the version based on T5-large uses a batch size of 8 with 30 epochs, both with a learning rate of 1e-4. For other baseline models, we carefully follow their official code for experiments.

We compare our method with following baselines, which are detailed below:
\begin{enumerate}
    \item \textbf{Sequential Fine-tuning} (\textbf{SFT}) \cite{yogatama2019learning} updates all model parameters for each incoming task in sequence, a technique proven to suffer from catastrophic forgetting. In our experiments, we maintain the same basic parameter settings as used in our approach to ensure consistency.
    \item \textbf{Rehearsal-based methods} avoid forgetting by repeatedly replaying real data from past tasks during the current task. Specifically, these methods typically allocate a memory space for each task to store past data and repeatedly replay this data in each training batch, directly benefiting continual learning tasks. In our experiments, the memory size for all replay-based methods is set to 10. It is important to note that our approach does not rely on replaying any real data from previous tasks, making this comparison not entirely fair and putting our approach at a disadvantage, which highlights its effectiveness.
    \begin{itemize}
        \item \textbf{EMAR} \cite{han-etal-2020-continual} utilizes episodic memory activation and reconsolidation to address catastrophic forgetting in continual learning, maintaining stable old relations while effectively incorporating new ones. 
        Note that we do not directly use the results from \citet{10.1609/aaai.v37i11.26492} and \citet{NEURIPS2023_398b00a0}. \citet{10.1609/aaai.v37i11.26492} uses GRAPPA\cite{yu2021grappa} as the backbone, which is pre-trained on specific table semantic parsing tasks. It generates a simplified intermediate representation instead of formal SQL, which cannot be executed on a database and loses accuracy when converted to SQL. To ensure fair and consistent comparisons and to avoid the confounding effects of specific pre-training tasks on the evaluation of continual learning (i.e., distinguishing whether the model's capabilities come from specific pre-training or continual learning), we uniformly adopt the T5 series models and evaluate the generated SQL using the official EM and EX metrics. 
        In \citet{NEURIPS2023_398b00a0}, since the task order for EMAR's three random runs on the Combined-stream is not specified and the memory size used differs from ours, so we cannot directly use the original results. To ensure a fair comparison, we carefully refer to the code from \citet{NEURIPS2023_398b00a0} and report the results under the current unified backbone and memory size settings, with other parameters following the original configurations.
        \item \textbf{SFNET} \cite{10.1609/aaai.v37i11.26492} combines semi-supervised and continual learning to tackle text-to-SQL tasks, using self-training and memory replay to enhance learning and maintain efficiency across tasks. 
        As mentioned in EMAR above, original SFNET also uses a specific pre-trained GRAPPA as the backbone and employs evaluation metrics tailored to simplified intermediate representations. We fairly use the T5 model as the backbone, set the warm start epoch to 50 for the base version and 30 for the large version, and evaluate the final SQL output using official metrics. 
        Notably, SFNET can leverage additional unlabeled data for unsupervised learning, allowing it to function without data replay, unlike EMAR. Therefore, we report SFNET's results under both replay and no-replay settings. In contrast, our method neither relies on replay nor requires additional unlabeled data.
    \end{itemize}

    \item \textbf{PET-based methods} first require a large amount of labeled data to warm up the initial model, and then learn a PET module for each task through efficient parameter fine-tuning methods such as prompt tuning. During inference, these methods typically operate under two settings:
    \textbf{One} that relies on an ideal setting, such as methods like PEFT and C3, which simplifies the continual learning scenario. Specifically, they assume the task identity for each test sample is known and load the corresponding pre-trained PET module. However, this setting is unrealistic in practice and cannot generalize to unseen samples, as they do not belong to any prior tasks, making appropriate module selection impossible.
    \textbf{The other} is the same general setting as our method, such as ProgPrompt. These methods do not require replaying any historical data and do not rely on ideal settings, making them more adaptable to real-world scenarios.
    \begin{itemize}
        \item \textbf{PEFT} \cite{NEURIPS2023_398b00a0} freezes most parameters in the backbone model and trains a unique prefix prompt for each task, effectively minimizing catastrophic forgetting caused by parameter updates. On the Combined-stream, we report the results from the original paper (PEFT's results are task order-independent). On the Spider-stream-semi, we reproduce the results using the original code. We set the prompt length to 150, prompt learning rate to 0.3, evaluation delay to 200, evaluation interval to 1, and early stopping patience to 5.
        \item \textbf{C3} \cite{NEURIPS2023_398b00a0} builds on PEFT by adopting an in-context tuning method,  where it leverages a few training examples to gather contextual information, and introduces a teacher-student framework to address the few-shot problem. Since the original paper does not specify the task order for the three random runs on the Combined-stream, we report the results from their code repository using the same order as ours. 
        \item \textbf{ProgPrompt} \cite{razdaibiedina2023progressive} addresses the need for additional task labels and the limitations on forward transfer in the PEFT method by constructing progressive prompts. Considering the ProgPrompt is originally designed for classification tasks, we carefully adjust its original code to adapt it for CSP tasks. As ProgPrompt uses a progressively increasing prompt design, the prompt length grows with the number of tasks, which may lead to truncation of the original input sequence and negatively impact performance. Therefore, we set the prompt length to 50, prompt learning rate to 0.3, evaluation delay to 150, evaluation interval to 2, and early stopping patience to 5.
    \end{itemize}

    \item \textbf{ORACLE} represents the theoretical upper bound of continual learning performance. It trains on all data seen so far for each task, so the training data increases as the number of tasks grows. The basic parameters are consistent with our method.

    \item \textbf{Different LLMs as Teacher 1} The LLMs we use include: {Mixtral-8x7B-Instruct-v0.1}\footnote{https://huggingface.co/mistralai/Mixtral-8x7B-Instruct-v0.1}, {Meta-Llama-3-8B-Instruct}\footnote{https://huggingface.co/meta-llama/Meta-Llama-3-8B-Instruct}, {Qwen2-72B-Instruct}\footnote{https://huggingface.co/Qwen/Qwen2-72B-Instruct}. 
    All model inferences use the vLLM\footnote{https://github.com/vllm-project/vllm} framework with the temperature at 0.6, top\_p at 0.95, and max\_tokens at 150, other parameters are set to default values.
    
\end{enumerate}

\begin{table*}[htbp]
\centering
\resizebox{\textwidth}{!}{%
\begin{tabular}{lccccccccc}   %
\toprule

\multirow{2}{*}{\textbf{Method}} & \multicolumn{1}{c}{\multirow{2}{*}{\textbf{DR}}} & \multicolumn{4}{c}{\textbf{Spider-stream-semi (\emph{cold start})}} & \multicolumn{4}{c}{\textbf{Combined-stream (\emph{cold start})}} \\
\cmidrule(lr){3-10} 
 &  & \multicolumn{1}{c}{$\text{ACC}_\text{a}$} & \multicolumn{1}{c}{$\text{ACC}_\text{w}$} & \multicolumn{1}{c}{BWT} & \multicolumn{1}{c}{FWT} & \multicolumn{1}{c}{$\text{ACC}_\text{a}$} & \multicolumn{1}{c}{$\text{ACC}_\text{w}$} & \multicolumn{1}{c}{BWT} & \multicolumn{1}{c}{FWT} \\ 
 \hline
\rowcolor{gray!30} \multicolumn{10}{c}{\textbf{\emph{T5-Base(220M)}}} \\
SFT & {\ding{55}} & 40.2/42.7 & 42.3/44.4 & -10.8/-9.8 & \underline{19.1}/\underline{20.9} & 39.9/36.4 & 32.0/29.4 & -25.9/-28.5 & 24.5/\underline{20.5}\\
ProgPrompt & {\ding{55}} & 0.0/0.3 & 0.0/0.3 & -3.8/-5.6 & 1.6/1.6 & 0.2/0.6 & 0.1/0.3 & -10.2/-11.2 & 0.1/0.0 \\
EMAR & {\ding{51}} & \underline{46.4}/\underline{48.3} & \underline{47.6}/\underline{49.5} & \underline{-2.1}/-2.4 & 17.1/19.0 & \underline{55.0}/51.9 & 51.6/50.6 & -10.2/-11.9 & 24.6/19.7 \\
SFNET$^{\clubsuit}$  & {\ding{55}} & 39.9/42.4 & 42.0/44.6 & -3.6/-2.9 & 18.1/19.8 & 34.5/31.9 & 24.4/23.3 & -33.7/-34.6 & 22.5/19.2 \\
SFNET$^{\clubsuit}$ & {\ding{51}} & {42.4}/{44.7} & 44.8/46.9 & {-2.5}/\underline{-2.2} & {18.2}/{20.0} & 54.0/\underline{52.3} & \underline{51.8}/\underline{51.5} & \underline{-8.8}/\underline{-9.3} & \underline{25.4}/20.4  \\
\textsc{Lecsp}(Ours) & {\ding{55}} & \textbf{48.8}/\textbf{52.6} & \textbf{49.9}/\textbf{54.2} & \textbf{-1.1}/\textbf{-0.9} & \textbf{24.8}/\textbf{27.3} & \textbf{58.6}/\textbf{56.8} & \textbf{57.6}/\textbf{56.8} & \textbf{-4.6}/\textbf{-4.3} & \textbf{29.6}/\textbf{29.2} \\
\hline
\multicolumn{10}{c}{\emph{Ideal Setting \& Upper Bound}} \\
PEFT$^{\diamondsuit}$ & {\ding{55}} & 8.1/11.6 & {10.0}/{14.5} & 0.0/0.0 & -/- & 25.0/23.1 & 16.6/18.2 & 0.0/0.0 & -/- \\
ORACLE & {} & 59.2/60.9 & 61.2/63.2 & 10.9/10.3 & 20.1/22.6 & 70.5/67.9 & 71.5/70.4 & 10.1/8.7 & 25.8/23.4 \\
\hline\hline
\rowcolor{gray!30} \multicolumn{10}{c}{\textbf{\emph{T5-Large(770M)}}} \\
SFT & {\ding{55}} & {48.2}/{50.2} & 49.5/51.8 & -6.0/-6.2 & \underline{26.9}/\underline{29.5} & 49.5/47.2 & 44.3/41.7 & -22.1/-23.0 & \underline{32.6}/27.0\\

ProgPrompt & {\ding{55}} & 1.3/1.7 & 0.9/1.3 & -3.6/-4.0 & 2.2/2.6 & 1.1/1.5 & 1.4/1.5 & -13.5/-14.1 & 0.3/0.2 \\
EMAR & {\ding{51}} & \underline{50.3}/\underline{52.8} & 51.9/\underline{54.4} & \underline{-3.6}/\underline{-2.3} & 25.9/27.6 & 58.4/56.6 & 57.8/57.0 & \textbf{-2.3}/\textbf{-2.9} & 28.7/22.5 \\
SFNET$^{\clubsuit}$ & {\ding{55}} & 47.5/50.1 & 49.2/51.8 & -4.3/-3.6 & 26.8/29.2 & 40.9/37.0 & 32.6/29.7 & -29.9/-33.5 & 28.8/25.0 \\
SFNET$^{\clubsuit}$ & {\ding{51}} & {50.0}/{51.7} & \underline{52.0}/53.9 & {-3.7}/\textbf{-2.7} & {24.9}/{27.0} & \underline{59.7}/\underline{58.7} & \underline{58.7}/\underline{58.6} & -8.3/-8.9 & 32.1/\underline{27.1} \\ 
\textsc{Lecsp}(Ours) & {\ding{55}} & \textbf{57.8}/\textbf{62.1} & \textbf{58.4}/\textbf{63.3} & \textbf{-2.2}/\underline{-1.1} & \textbf{35.9}/\textbf{38.4} & \textbf{63.1}/\textbf{61.4} & \textbf{63.4}/\textbf{62.1} & \underline{-3.5}/\underline{-4.0} & \textbf{35.2}/\textbf{32.7} \\
\hline
\multicolumn{10}{c}{\emph{Ideal Setting \& Upper Bound}} \\
PEFT$^{\diamondsuit}$ & {\ding{55}} & 25.5/29.3 & 29.3/33.7 & 0.0/0.0 & -/- & 43.9/45.5 & 38.2/40.8 & 0.0/0.0 & -/- \\
ORACLE & {} & 67.0/68.3 & 68.5/70.2 & 11.0/10.9 & 26.6/29.4 & 73.6/73.2 & 75.0/75.6 & 5.4/5.7 & 37.7/34.4 \\
\bottomrule
\end{tabular}
}
\caption{Results (EM/EX) on Spider-stream-semi and Combined-stream datasets under \emph{cold start} settings (\%). 
\textsc{Lecsp} uses Mixtral-8x7B as \emph{Teacher 1}. DR indicates whether historical data replay is used, with a memory size set to 10. 
$\clubsuit$ indicates the need for additional unsupervised data, and ${\diamondsuit}$ means an ideal continuous learning setting is used, but forward transfer is impossible. The best results are highlighted in bold, and the second-best results are underlined. Our results are the average of three random runs.}
\label{tab:result_cold}
\end{table*}

\begin{table*}[hbp]
\centering
\resizebox{\textwidth}{!}{%
\begin{tabular}{lccccccccc}   %
\toprule

\multirow{2}{*}{\textbf{Method}} & \multicolumn{1}{c}{\multirow{2}{*}{\shortstack{\textbf{Memory} \\ \textbf{Size}}}} & \multicolumn{4}{c}{\textbf{Spider-stream-semi}} & \multicolumn{4}{c}{\textbf{Combined-stream}} \\
\cmidrule(lr){3-10} 
 &  & \multicolumn{1}{c}{$\text{ACC}_\text{a}$} & \multicolumn{1}{c}{$\text{ACC}_\text{w}$} & \multicolumn{1}{c}{BWT} & \multicolumn{1}{c}{FWT} & \multicolumn{1}{c}{$\text{ACC}_\text{a}$} & \multicolumn{1}{c}{$\text{ACC}_\text{w}$} & \multicolumn{1}{c}{BWT} & \multicolumn{1}{c}{FWT} \\ 
 \hline

\multicolumn{10}{c}{{\emph{Original Method}}} \\
SFT & {0} & 41.7/45.3 & 40.1/43.7 & -13.6/-11.7 & 22.5/25.2 & 51.0/36.0 &  52.4/38.3 & -11.1/-17.6 & 19.9/11.1\\
\hline
\multicolumn{10}{c}{{\emph{Rehearsal-Based Method}}} \\
\multirow{3}{*}{EMAR} & {1} & 39.2/43.5 & 37.9/42.5 & -16.5/-13.6 & 23.6/25.3 & 53.7/51.2 & 50.8/49.4 & -15.8/-16.5 & 23.1/19.0 \\
{} & {10} & 44.1/48.0 & 43.0/47.0 & -11.8/-9.7 & 23.7/25.2 & 60.6/58.9 & 55.7/55.3 & -7.9/-7.1 & 28.7/25.4  \\
{} & {15} & 47.7/50.6 & 46.1/49.4 & -7.8/-7.5 & 23.5/25.4 & 61.2/59.4 & 57.6/57.9 & -7.3/-6.8 & 28.4/25.9 \\
\hline
\multicolumn{10}{c}{{\emph{Rehearsal-Based Method Requiring Unsupervised Data}}} \\
\multirow{4}{*}{SFNET}  & {0} & 39.5/43.9 & 38.1/42.7 & -14.6/-12.4 & 22.7/25.7 & 56.2/49.4 &  51.8/48.4 &  -7.9/-14.4 & 21.0/14.6 \\
{} & {1} & 42.0/45.8 & 40.4/44.6 & -11.2/-7.8 & 24.1/26.6 & 59.1/54.1 & 53.3/50.8 & -6.3/-9.7 & 25.6/19.9 \\
{} & {10} & 45.8/48.1 & 44.2/47.7 & -6.4/-6.4 & 23.9/26.2 & 60.7/57.7 & 55.2/53.9 &  -6.0/-6.7 & 28.7/26.0  \\
{} & {15} & \textbf{47.9}/51.2 & \textbf{46.8}/49.9 & -6.0/-5.8 & 25.0/27.1 & 61.3/60.9 & 57.5/56.7 & -5.4/-5.2 & 29.7/27.5  \\
\hline

\multicolumn{10}{c}{{\emph{Without Rehearsal and Unsupervised Data}}} \\
\textsc{Lecsp}(Ours) & {0} & 47.4/\textbf{53.7} & 46.5/\textbf{53.3} & \textbf{-4.7}/\textbf{-4.6} & \textbf{30.1}/\textbf{32.8} & \textbf{63.4}/\textbf{61.7} &  \textbf{60.5}/\textbf{59.8} & \textbf{-4.6}/\textbf{4.3} & \textbf{32.7}/\textbf{31.5} \\
\hline
\multicolumn{10}{c}{\emph{Upper Bound}} \\
ORACLE & {} & 60.3/61.3 & 62.8/63.9 &  4.6/3.3 & 25.7/27.9 & 70.2/68.2 & 71.1/70.8 &  5.6/4.7 &  29.2/26.6  \\
\bottomrule
\end{tabular}
}
\caption{Comparison (EM/EX) of our method with rehearsal-based methods using different memory sizes (\%).
\textsc{Lecsp} uses Mixtral-8x7B as \emph{Teacher 1}. 
The best results are highlighted in bold. Our results are the average of three random runs.}
\label{tab:result_memory_size}
\end{table*}

\section{C\quad More Results and Analysis}
\label{apx:more_results}

\subsection{C.1\quad Results under Cold Start Settings and Different Memory Sizes}
\label{apx:cold_start}

Considering the warm start settings of previous works \cite{10.1609/aaai.v37i11.26492, NEURIPS2023_398b00a0}, we explore a more challenging and realistic cold start scenario. As shown in Table \ref{tab:result_cold}, the performance of most methods declined, particularly for \emph{PET-based} methods like PEFT and ProgPrompt. This aligns with our expectations, as these methods can no longer rely on the data-rich initial task for good initialization parameters. However, \textsc{Lecsp} demonstrates better robustness. 

Table \ref{tab:result_memory_size} also compares our method with \emph{Rehearsal-based} methods across different memory sizes. The results show that the performance of \emph{Rehearsal-based} methods is highly dependent on memory size. For EMAR, when the memory size is reduced to 1, its performance drops significantly across several metrics, even falling below that of SFT. 
While SFNET can function without replaying previous data by utilizing additional unlabeled data for semi-supervised learning (memory size = 0), its performance remains poor and is still constrained by memory size. In contrast, our method consistently delivers excellent performance without the need for data replay or unlabeled data.

\begin{table}[htbp]
\centering
\resizebox{0.48\textwidth}{!}{%
\begin{tabular}{lcccc}
\toprule
{\textbf{Model}} & \multicolumn{1}{c}{$\text{ACC}_{\text{a}}$} & \multicolumn{1}{c}{$\text{ACC}_{\text{w}}$} & \multicolumn{1}{c}{BWT} & \multicolumn{1}{c}{FWT} \\ 
\hline
{SFT} & 41.7/45.3 & 40.1/43.7 & -13.6/-11.7 & 22.5/25.2 \\
\hline
{Mixtral-8x7B} & 22.0/49.4 & 22.2/52.4 & -/- & -/- \\
+{\textsc{Lecsp}}  & 47.4/53.7  & 46.5/53.3  & -4.7/-4.6  & 30.1/32.8 \\
\hline
{Llama-3-8B} & 16.9/56.3 & 16.6/59.2 & -/- & -/- \\
+{\textsc{Lecsp}}  & 49.7/54.7  & 47.6/53.3  & -8.3/-6.9  & 31.0/33.4 \\
\hline
{Qwen-2-72B} & 45.6/49.9 & 45.5/49.2 & -/- & -/- \\
+{\textsc{Lecsp}}  & 47.2/49.8  & 45.7/50.4  & -10.0/-9.0  & 29.5/31.6 \\
\hline
ORACLE & 60.3/61.3 & 62.8/63.9 & 4.6/3.3 & 25.7/27.9 \\

\bottomrule
\end{tabular}
}
\caption{Zero-shot results of LLMs and their performance as \emph{Teacher 1} for guiding smaller models (T5-base) on Spider-stream-semi dataset (\%).}
\label{tab:diff_teacher}
\end{table}

\subsection{C.2\quad Impact of Different LLMs as {Teacher 1}}
\label{apx:diff_teacher}

For \emph{Teacher 1}, we use three different series and scales of LLMs to explore their impact on performance. As shown in Table \ref{tab:diff_teacher}, all three LLMs effectively improve the continual learning performance of smaller models in terms of $\text{ACC}_\text{a}$ and $\text{ACC}_\text{w}$, with Mixtral-8x7B performing the best on the BWT metric. Notably, with the introduction of LLM teachers, the FWT performance of the student models consistently surpasses the upper bound.

\subsection{C.3\quad Performance Comparison Across Different Difficulty Levels}

\begin{figure*}[ht]
\centering
\includegraphics[width=0.7\textwidth]{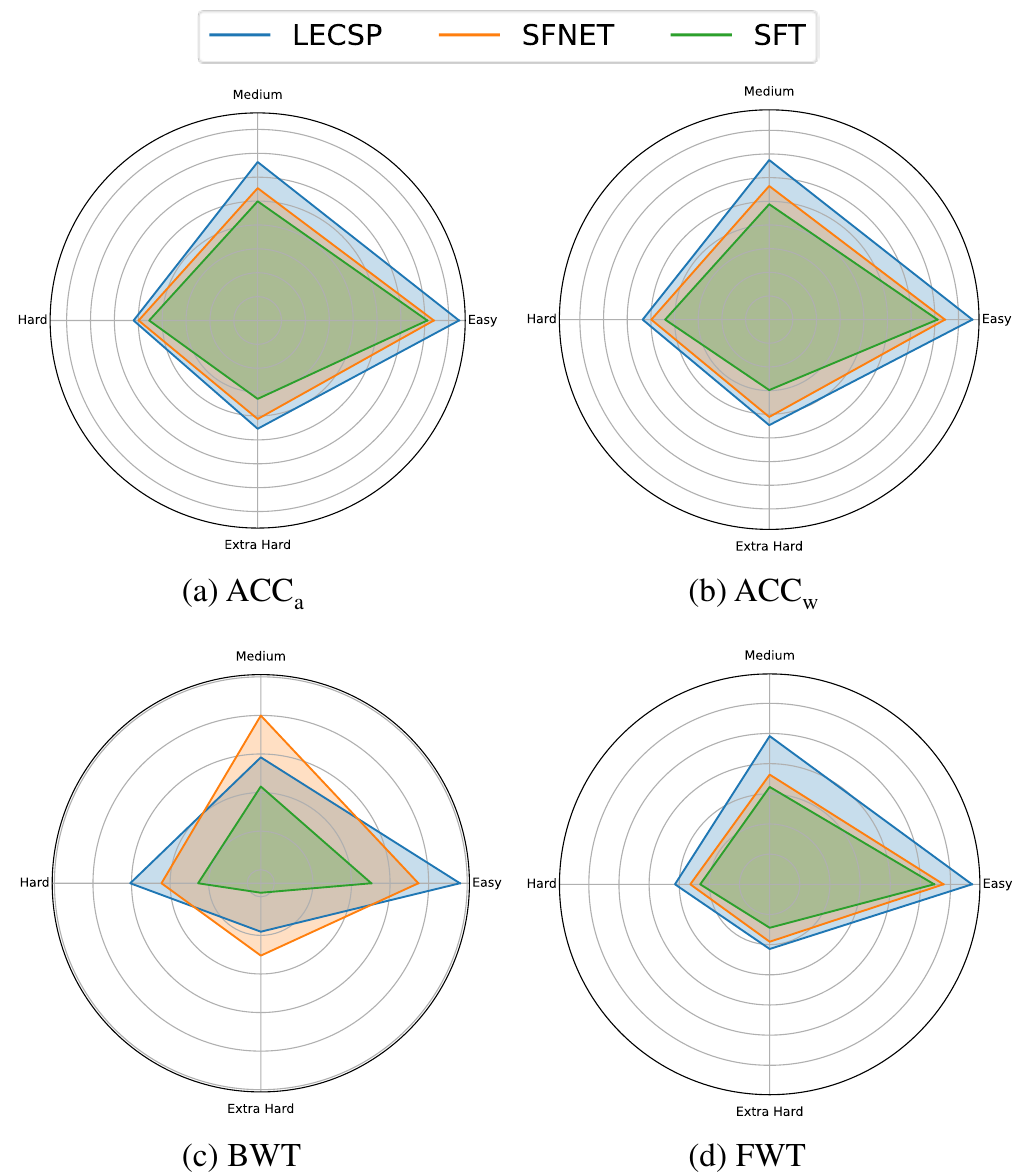}
\caption{Performance (EX) comparison at different difficulty levels on Spider-stream-semi dataset (T5-large).} 
\label{fig:radar_ex_large}
\end{figure*}

Figure~\ref{fig:radar_ex_large} presents a comparative analysis of method performances across difficulty levels including easy, medium, hard, and extra hard. The results demonstrate that our method consistently outperforms the baseline across almost all metrics, particularly showing notable improvements at easy and medium difficulties. Regarding the BWT metric, our approach performs comparably to SFNET. This can be attributed to SFNET's data replay strategy, which inherently mitigates the negative impact of training on new tasks on previous tasks. In contrast, our method, which reconstructs memory using an external LLM, might introduce new knowledge.

\subsection{C.4\quad Results Till the Seen Tasks}
\label{apx:seen_task}

\begin{figure*}[ht]
\centering
\includegraphics[width=0.8\textwidth]{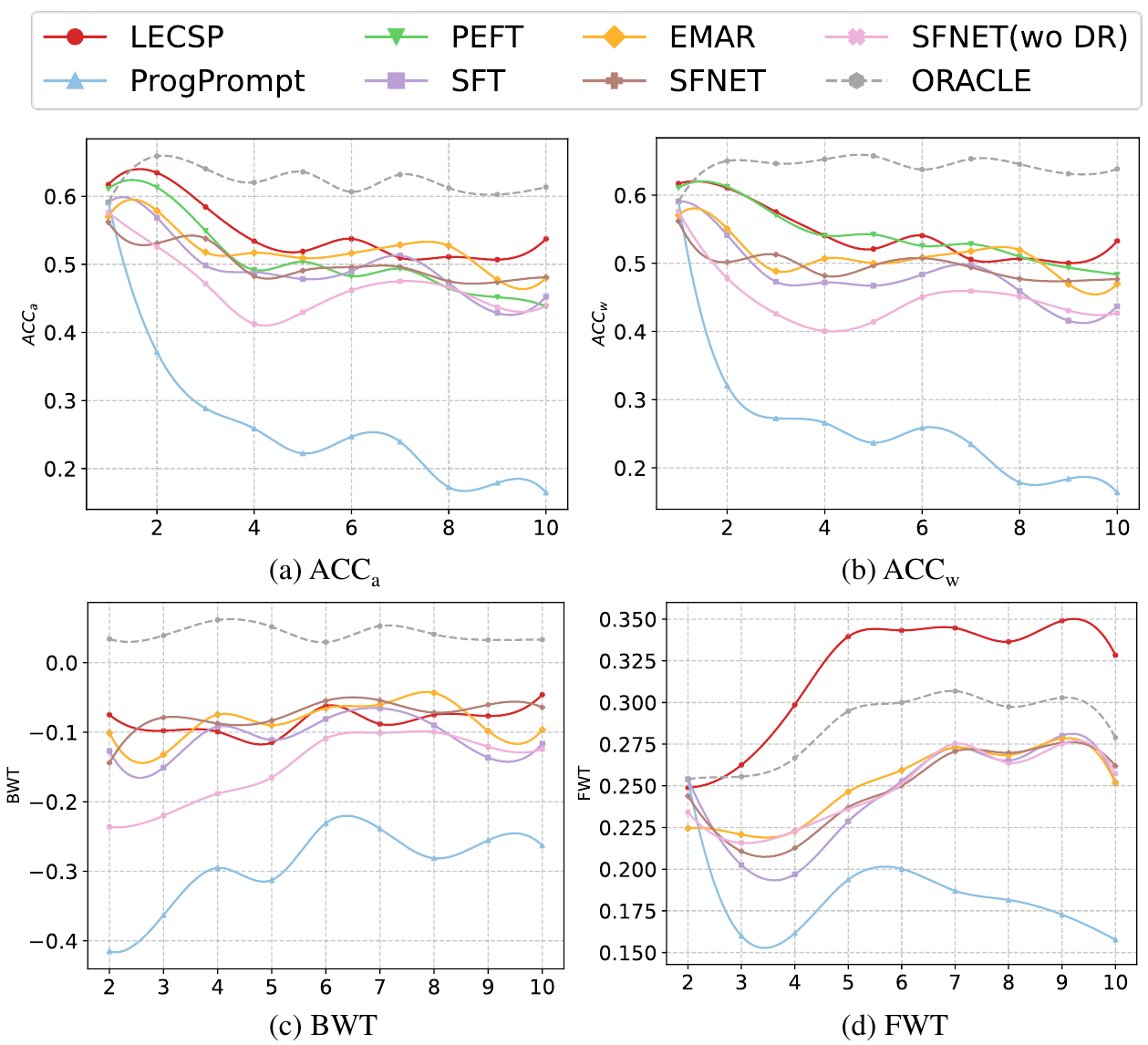}
\caption{Results (EX) till the seen tasks on Spider-stream-semi dataset (T5-base).
} 
\label{fig:result_line_spider_base}
\end{figure*}

\begin{figure*}[ht]
\centering
\includegraphics[width=0.8\textwidth]{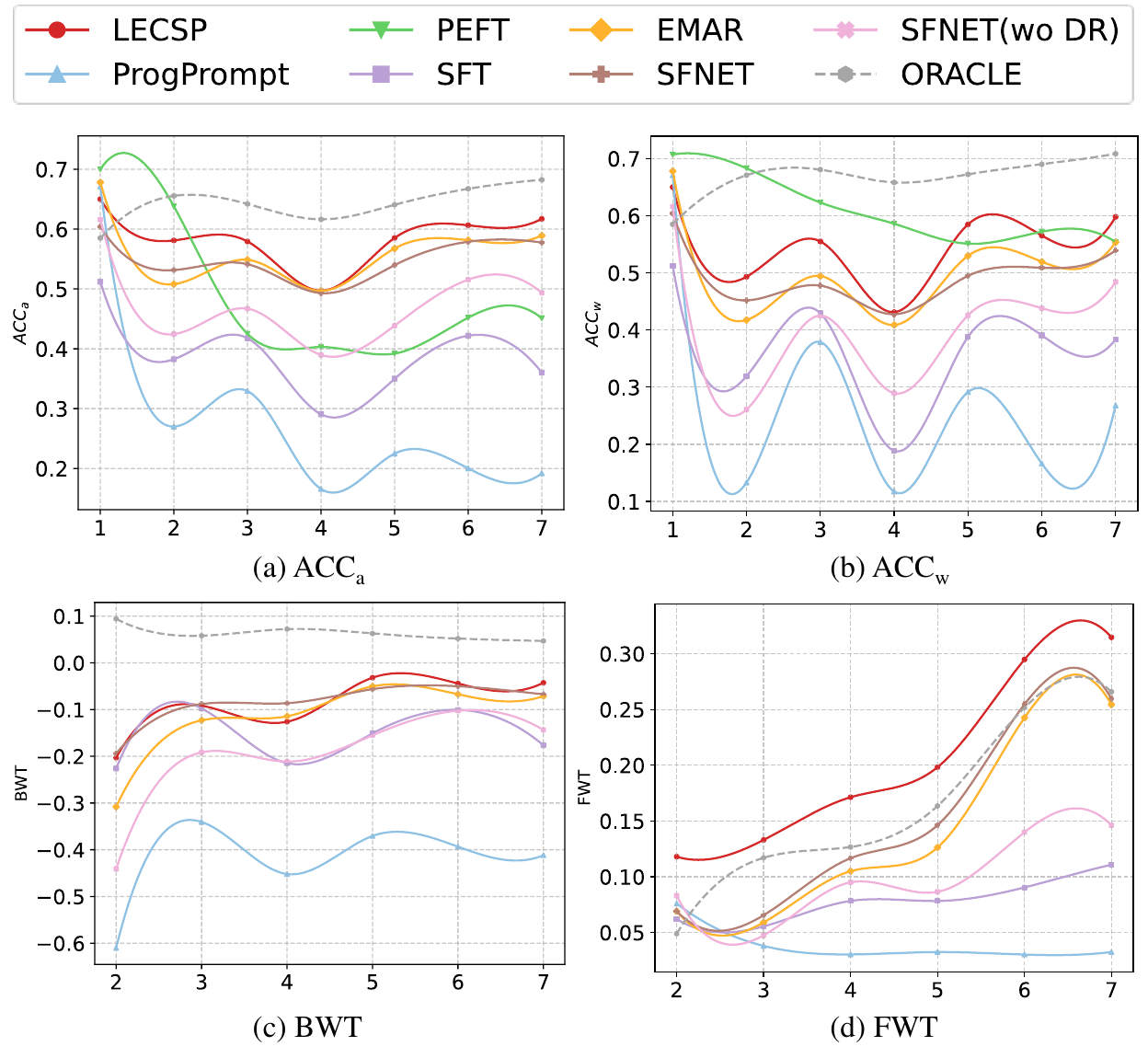}
\caption{Results (EX) till the seen tasks on Combined-stream dataset (T5-base).
} 
\label{fig:result_line_combine_base}
\end{figure*}

\begin{figure*}[ht]
\centering
\includegraphics[width=0.8\textwidth]{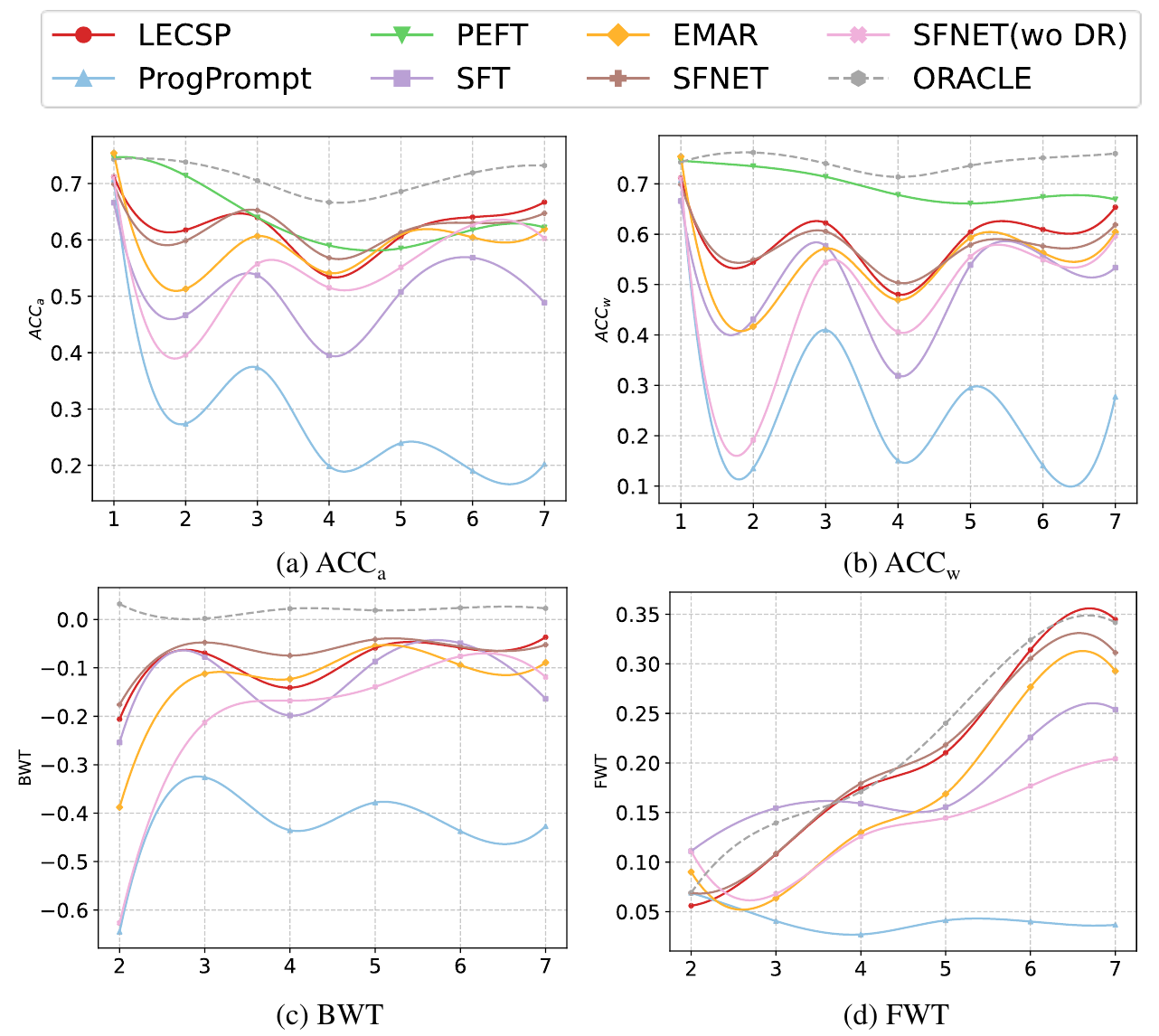}
\caption{Results (EX) till the seen tasks on Combined-stream dataset (T5-large).
} 
\label{fig:result_line_combine_large}
\end{figure*}

Figure~\ref{fig:result_line_spider_base} to Figure~\ref{fig:result_line_combine_large} shows more results of different methods on previously seen tasks. Our method continues to maintain a stable overall performance advantage.

\begin{table}[htbp]
\centering
\resizebox{0.48\textwidth}{!}{%
\begin{tabular}{lcccccc}
\toprule
{} & {} & \multicolumn{1}{c}{$\text{ACC}_{\text{a}}$} & \multicolumn{1}{c}{$\text{ACC}_{\text{w}}$} & \multicolumn{1}{c}{BWT} & \multicolumn{1}{c}{FWT} \\ 
\hline
\multicolumn{2}{l}{\textsc{Lecsp} } & \textbf{47.4}/\textbf{53.7}  & \textbf{46.5}/\textbf{53.3}  & -4.7/\textbf{-4.6}  & 30.1/32.8 \\
\multicolumn{2}{c}{--\emph{only} $\mathcal{Z}$} & 43.4/49.1 & 42.4/48.4 & -4.2/-5.1 & 26.5/29.0 \\
\hline
\multirow{3}{*}{$K$} & 60 & 44.2/49.4 & 43.0/48.7 & \textbf{-4.0}/-4.9 & 26.5/29.4 \\
& 70 & 46.1/52.4 & 45.5/52.3 & -5.9/-6.1 & 29.5/32.6 \\ 
& 90 & 46.9/53.3 & 46.4/53.0 & -7.3/-6.1 & 30.6/33.1 \\ 
\hline
\multirow{4}{*}{$\lambda$} & 0.03 & 46.7/53.2 & 45.9/52.9 & -5.8/-5.5 & 30.8/\textbf{33.3} \\
& 0.05 & 46.9/53.3 & 46.1/53.1 & -5.7/-5.3 & \textbf{30.9}/\textbf{33.3} \\ 
& 0.2 & 46.7/52.8 & 46.3/52.9 & -6.9/-6.2 & 30.4/33.0 \\
& 0.3 & 47.0/53.0 & {46.4}/52.9 & -6.0/-5.6 & 30.6/33.2 \\
\bottomrule
\end{tabular}
}
\caption{Ablation study results of hyperparameter and clustering settings on the Spider-stream-semi (\%). In \textsc{Lecsp}, $K$=80, $\lambda$=0.1, and both the domain-agnostic question ($\mathcal{Q}^{de}$) and SQL skeleton ($\mathcal{Z}$) are used as the basis for clustering.}
\label{tab:lambda}
\end{table}

\subsection{C.5\quad Additional Ablation Study Results}
\label{apx:lambda}

Table \ref{tab:lambda} provides detailed results of the hyperparameter experiments. The hyperparameter $\lambda$ significantly affects BWT, reaching up to 2.2\%, with optimal overall performance achieved at 0.1. 
Regarding the hyperparameter $K$, we find that a low number of cluster centers reduces the component features $\mathcal{A}$ obtained for each task, affecting the induction of component bias $\Delta\mathcal{A}$ and leading to a decrease in model performance. As $K$ increases, the growth of the component feature set size gradually slows, as some central samples have redundant SQL skeletons, reducing the model's sensitivity to the number of cluster centers.

We also explore an alternative clustering setup, where only the SQL skeleton ($\mathcal{Z}$) is used as the basis for clustering, without combining the domain-agnostic question ($\mathcal{Q}^{de}$) and SQL skeleton ($\mathcal{Z}$). The results in Table \ref{tab:lambda} show that relying solely on the SQL skeleton ($\mathcal{Z}$) leads to a general decline in performance. This may be because in the semantic parsing task, the same question can correspond to different SQL implementations, and the same SQL can be expressed in multiple ways. Therefore, relying only on $\mathcal{Z}$ may fail to capture the full diversity and complexity of samples. Considering both the question and SQL helps better represent and distinguish samples during clustering, leading to more accurate component bias analysis.




\end{document}